\documentclass{article}

\usepackage{microtype}
\usepackage{graphicx}
\usepackage{booktabs} % for professional tables

\usepackage{amsfonts}       % blackboard math symbols
\usepackage{amsmath}       % blackboard math symbols

\usepackage{tikz}
\usetikzlibrary{fadings}

\usepackage{wasysym}

\usepackage{subcaption}
\usepackage{epigraph}
\usepackage{booktabs}
\usepackage{colortbl}

\newcommand{\yourvalue}{\mathbf{v}^i}
\newcommand{\yourallocation}{X^i}
\newcommand{\yourpayment}{P^i}
\newcommand{\yourutility}{U^i}
\newcommand{\winnerutility}{\hat{U}^i}

\newcommand{\allbidders}{\psi^{1:N}}
\newcommand{\allprimitives}{\omega^{1:N}}

\newcommand{\yourindex}{i}

\newcommand{\lastagentindex}{N}

\newcommand{\nextindex}{k}

\usepackage{amsthm}
\newtheorem{definition}{Definition}[section]
\newtheorem{theorem}{Theorem}[section]

\newtheorem{lemma}[theorem]{Lemma}
\newtheorem{proposition}[theorem]{Proposition}

\theoremstyle{remark}

\usepackage{algorithm}
\usepackage{algpseudocode}

\algnewcommand{\LineComment}[1]{\State \(\triangleright\) #1}

\newcommand{\primitive}{\omega}

\newcommand{\agentbidder}{\psi}

\newcommand{\agenttransformation}{\phi}
\newcommand{\yourprimitive}{\primitive^i}
\newcommand{\winnerprimitive}{\hat{\primitive}}

\DeclareCaptionLabelFormat{andtable}{#1~#2  \&  \tablename~\thetable}

\usepackage[export]{adjustbox}

\newcommand{\bucketbrigade}{\emph{BB} }

\newcommand{\societalQ}{Q_\Omega}

\usepackage{dblfloatfix} 

\usepackage{amsmath,amsfonts,bm}

\def\eqref#1{equation~\ref{#1}}
\def\1{\bm{1}}

\DeclareMathAlphabet{\mathsfit}{\encodingdefault}{\sfdefault}{m}{sl}
\SetMathAlphabet{\mathsfit}{bold}{\encodingdefault}{\sfdefault}{bx}{n}

\DeclareMathOperator*{\argmax}{arg\,max}

\usepackage{hyperref}

\usepackage[accepted]{icml2020}

\icmltitlerunning{Decentralized Reinforcement Learning: Global Decision-Making via Local Economic Transactions}

\begin{document}

\twocolumn[
\icmltitle{Decentralized Reinforcement Learning:\\Global Decision-Making via Local Economic Transactions}



\icmlsetsymbol{equal}{*}

\begin{icmlauthorlist}
\icmlauthor{Michael Chang}{1}
\icmlauthor{Sidhant Kaushik}{1}
\icmlauthor{S. Matthew Weinberg}{2}
\icmlauthor{Thomas L. Griffiths}{2}
\icmlauthor{Sergey Levine}{1}
\end{icmlauthorlist}

\icmlaffiliation{1}{Department of Computer Science, University of California, Berkeley, USA}
\icmlaffiliation{2}{Department of Computer Science, Princeton University, USA}

\icmlcorrespondingauthor{Michael Chang}{mbchang@berkeley.edu}

\icmlkeywords{Machine Learning, ICML}

\vskip 0.3in
]



\printAffiliationsAndNotice{}  % leave blank if no need to mention equal contribution

\begin{abstract}
This paper\footnote{Code, talk, and blog \href{https://sites.google.com/view/clonedvickreysociety/home}{\textcolor{blue}{here}}.} seeks to establish a framework for directing a society of simple, specialized, self-interested agents to solve what traditionally are posed as monolithic single-agent sequential decision problems. What makes it challenging to use a decentralized approach to collectively optimize a central objective is the difficulty in characterizing the equilibrium strategy profile of non-cooperative games. To overcome this challenge, we design a mechanism for defining the learning environment of each agent for which we know that the optimal solution for the global objective coincides with a Nash equilibrium strategy profile of the agents optimizing their own local objectives. The society functions as an economy of agents that learn the credit assignment process itself by buying and selling to each other the right to operate on the environment state. We derive a class of decentralized reinforcement learning algorithms that are broadly applicable not only to standard reinforcement learning but also for selecting options in semi-MDPs and dynamically composing computation graphs. Lastly, we demonstrate the potential advantages of a society's inherent modular structure for more efficient transfer learning.
\end{abstract}

\epigraph{You know that everything you think and do is thought and done by you. But what's a ``you''? What kinds of smaller entities cooperate inside your mind to do your work?}{\citep{minsky1988society}}

\begin{figure}
    \centering
    \includegraphics[width=0.4\textwidth]{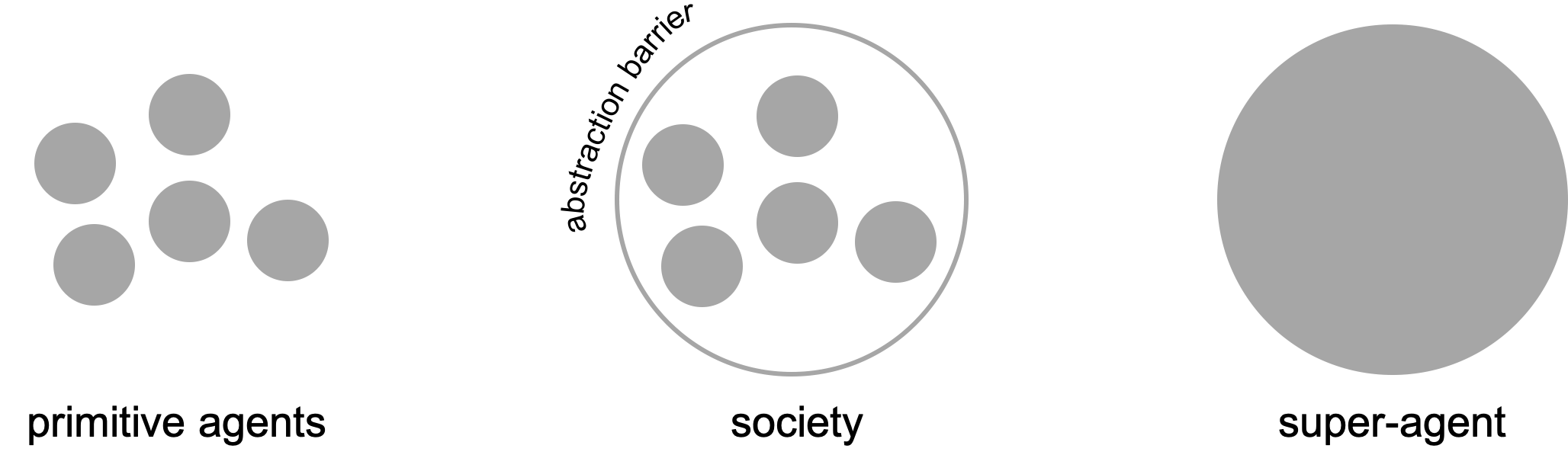}
    \caption{\small{
    We study how a society of primitives agents can be abstracted as a super-agent.
    The incentive mechanism is the abstraction barrier that relates the optimization problems of the super-agent with those of its constituent primitive agents.
    }}
    \label{fig:main}
\end{figure}

\section{Introduction}
Biological processes, corporations, and ecosystems -- physically decentralized, yet in some sense functionally unified.
A corporation, for example, optimizes for maximizing profits as it were a single rational agent.
But this agent abstraction is an illusion: the corporation is simply a collection of human agents, each solving their own optimization problems, most not even knowing the existence of many of their colleagues.
But the human as the decision-making agent is also simply an abstraction of the trillions of cells making their own simpler decisions.
\textit{The society of agents is itself an agent.}
What mechanisms bridge between these two levels of abstraction, and under what framework can we develop learning algorithms for studying the self-organizing nature of intelligent societies that pervade so much of the world?

Both the monolithic and the multi-agent optimization frameworks in machine learning offer a language for representing only one of the levels of abstraction but not the relation between both.
The monolithic framework, the most commonly used in much of modern machine learning, considers a single agent that optimizes a single objective in an environment, whether it be minimizing classification loss or maximizing return.
The multi-agent framework considers multiple agents that each optimize their own independent objective and each constitute each other's learning environments.
What distinguishes the multi-agent from the monolithic is the presence of multiple independent optimization problems.
The difficulty of interpreting a learner in the monolithic framework as a society of simpler components is that all components are still globally coupled together by the same optimization problem without independent local optimization problems themselves, as are the weights in a neural network trained by backpropagation.
The difficulty of interpreting a multi-agent system under a global optimization problem is the computational difficulty of computing Nash equilibrium~\citep{daskalakis2009complexity}, even for general two-player games~\citep{chen2009settling}.

To better understand the relationship between the society and the agent, this paper makes four contributions, each at a different level of abstraction.
At the highest level, we define the \textbf{societal decision-making} \textit{framework} to relate the local optimization problem of the agent to the global optimization problem of the society in the following restricted setting.
Each agent is specialized to transform the environment from one state to another.
The agents bid in an auction at each state and the auction winner transforms the state into another state, which it sells to the agents at the next time-step, thereby propagating a series of economic transactions.
This framework allows us to ask what are properties of the auction mechanism and of the society that enable the global solution to a Markov decision process (MDP) that the society solves to emerge implicitly as a consequence of the agents optimizing their own independent auction utilities.

At the second level, we present a \textit{solution} to this question by introducing the \textbf{cloned Vickrey society} that guarantees that the dominant strategy equilibrium of the agents coincides with the optimal policy of the society.
We prove this result by leveraging the truthfulness property of the Vickrey auction~\citep{vickrey1961counterspeculation} and showing that initializing redundant agents makes the primitives' economic transactions robust against market bubbles and suboptimal equilibria.

At the third level, we propose a class of \textbf{decentralized reinforcement learning} \textit{algorithms} for optimizing the MDP objective of the society as an emergent consequence of the agents' optimizing their own auction utilities.
These algorithms treat the auction utility as optimization objectives themselves, thereby learning a societal policy that is global in space and time using only credit assignment for learnable parameters that is local in space and time.

At the fourth level, we empirically investigate various \textit{implementations} of the cloned Vickrey society under our decentralized reinforcement learning algorithm and find that a particular set of design choices, which we call the \textbf{credit conserving Vickrey} implementation, yields both the best performance at the societal and and agent level.

Finally, we demonstrate that the societal decision making framework, along with its solution, the algorithm that learns the solution, and the implementation of this algorithm, is a broadly applicable perspective on self-organization to not only standard reinforcement learning but also selecting options in semi-MDPs~\citep{sutton1999between} and composing functions in dynamic computation graphs~\citep{chang2018automatically}.
Moreover, we show evidence that the local credit assignment mechanisms of societal decision-making produce more efficient learning than the global credit assignment mechanisms of the monolithic framework.
\section{Related Work}
Describing an intelligent system as the product of interactions among many individual agents dates as far back as the Republic~\citep{plato380republic}, in which Plato analyzes the human mind via an analogy to a political state.
This theme continued into the early foundations of AI in the 1980s and 1990s through cognitive models such as the Society of Mind~\citep{minsky1988society} and Braitenberg vehicles~\citep{braitenberg1986vehicles} and engineering successes in robotics~\citep{brooks1991intelligence} and in visual pattern recognition~\citep{selfridge1988pandemonium}.

The closest works to ours were the algorithms developed around that same time period that sought as we do to leverage a multi-agent society for achieving a global objective, starting as early as the bucket brigade algorithm~\citep{holland1985properties}, in which agents bid in a first-price auction to operate on the state and auction winners directly paid their bid to the winners from the previous step.
Prototypical self-referential learning mechanisms~\citep{schmidhuber1987evolutionary} improved the bucket brigade by imposing credit conservation in the economic transactions.
The neural bucket brigade~\citep{schmidhuber1989local} adapted the bucket brigade to learning neural network weights, where payoffs corresponded to weight changes.
~\citet{baum1996toward} observed that the optimal choice for an agent's bid should be equivalent to the optimal Q-value for executing that agent's transformation and developed the Hayek architecture for introducing new agents and removing agents that have gone broke.
\citet{kwee2001market} added external memory to the Hayek architecture.

However, to this date there has been no proof to the best of our knowledge that the bid-updating schemes proposed in these works simultaneously optimize a global objective of the society in a decision-making context.
\citet{sutton1988learning} provides a convergence proof for temporal difference methods that share some properties with the bucket brigade credit assignment scheme, but importantly does not take the competition between the individual agents into account.
But it is precisely the competition among agents in multi-agent learning that make their equilibria nontrivial to characterize~\citep{mazumdar2019policy}.
Our work offers an alternative auction mechanism for which we prove that the optimal solution for the global objective \emph{does} coincide with a Nash equilibrium of the society.
We follow similar motivations to~\citet{balduzzi2014cortical}, which investigates incentive mechanisms for training a society of rational discrete-valued neurons.
In contrast to other works that decouple the computation graph~\citep{srivastava2013compete,goyal2019reinforcement,peng2019mcp,pathak2019learning} but optimize a global objective, our work considers optimizing local objectives only.
We consider economic transactions between time-steps, as opposed to within a single time-step~\citep{ohsawa2018neuron}.
\section{Preliminaries}
To set up a framework for societal decision-making, we relate Markov decision processes (MDP) and auctions under a unifying language.
We define an \textbf{environment} as a tuple that specifies an input space, an output space, and additional parameters for specifying an objective.
An \textbf{agent} is a function that maps the input space to the output space.
An \textbf{objective} is a functional that maps the learner to a real number.
Given an environment and objective, the \textbf{problem} the agent solves is to maximize the value of the objective.

In the MDP environment, the input space is the state space $\mathcal{S}$ and the output space is the action space $\mathcal{A}$.
The agent is a policy $\pi : \mathcal{S} \rightarrow \mathcal{A}$.
The transition function $\mathcal{T}: \mathcal{S} \times \mathcal{A} \rightarrow \mathcal{S}$, the reward function $r: \mathcal{S} \times \mathcal{A} \rightarrow \mathbb{R}$, and discount factor $\gamma$ are additional parameters that specify the objective: the return
$J(\pi) = \mathbb{E}_{\tau \sim p^\pi(\tau)}\left[\sum_{t=0}^T \gamma^t r\left(s_t, a_t\right)\right]$, where $p^\pi(\tau) = p(s_0)\prod_{t=0}^T \pi(a_t|s_t) \prod_{t=0}^{T-1} \mathcal{T}(s_{t+1}|s_t, a_t)$.
The agent solves the problem of finding $\pi^* = \argmax_\pi J(\pi)$.
For any state $s$, the optimal action for maximizing $J(\pi)$ is $\pi^*(s) = \argmax_a Q^*(s, a)$, where the optimal Q function $Q^*(s, a)$ is recursively defined as $Q^*(s, a) = \mathbb{E}_{s' \sim \mathcal{T}(s,a)}\left[r(s,a) + \gamma \max_{a'} Q^*(s', a') | s, a\right]$.

In the auction environments we consider, the input space is a single auction item $s$ and the output space is the bidding space $\mathcal{B}$.
Instead of a single agent, each of $N$ agents $\psi^{1:N}$ compete to bid for the auction item via its bidding policy $\psi^i: \{s\} \rightarrow \mathcal{B}$.
Let $\mathbf{b}$ be the vector of bids produced by $\psi^{1:N}$.
The vector $\mathbf{v}_s$ of each agent's valuations for auction item $s$ and the \textbf{auction mechanism} -- allocation rule $X: \mathcal{B}^N \rightarrow [0,1]^N$ and pricing rule $P: \mathcal{B}^N \rightarrow \mathbb{R}_{\geq 0}^N$ -- are additional parameters that specify each agent's objective: the utility $\yourutility_s(\allbidders) = \yourvalue_s \cdot \yourallocation(\mathbf{b}) - \yourpayment(\mathbf{b})$, where $\yourallocation(\mathbf{b})$ is the proportion of $s$ allocated to $i$, and $\yourpayment(\mathbf{b})$ is the scalar price $i$ pays.
Each agent $i$ independently solves the problem of finding $\psi^{i*} = \argmax_{\psi^{i}} \yourutility_s(\allbidders)$.
The independent optimization of objectives distinguishes a multi-agent problem from a single-agent one and makes multi-agent problems generally difficult to analyze when an agent's optimal policy depends on the strategies of other agents.

However, if an auction is dominant strategy incentive compatible (DSIC), bidding one's own valuation is optimal, independent of other players' bidding strategies. 
That is, truthful bidding is the unique dominant strategy.
Notably, the \textbf{Vickrey auction}~\citep{vickrey1961counterspeculation}, which sets $\yourpayment(\mathbf{b})$ to be the second highest bid $\max_{j \neq i} \mathbf{b}^j$ and $\yourallocation(\mathbf{b})=1$ if $i$ wins and $0$ and $0$ respectively if $i$ loses, is DSIC, which means the dominant strategy equilibrium occurs when every agent bids truthfully, making the Vickrey auction straightforward to analyze.
Another attractive property of the Vickrey auction is that the dominant strategy equilibrium automatically maximizes the social welfare $\sum_{i=1}^N \yourvalue \cdot \yourallocation(\mathbf{b})$~\citep{roughgarden2016twenty}, which selects the bidder with the highest valuation as winner.
The existence of dominant strategies in the Vickrey auction removes the need for agents to recursively model others, giving the Vickrey auction the practical benefit of running in linear time~\citep{roughgarden2016twenty}. 
\section{Societal Decision-Making}
The perspective of this paper is that a society of agents can be abstracted as an agent that itself solves an optimization problem at a global level as an emergent consequence of the optimization problems its constituent agents solve at the local level.
To make this abstraction precise, we now introduce the \textbf{societal decision-making} framework for analyzing and developing algorithms that relate the global decision problem of a society to the local decision problems of its constituent agents.
We use \textbf{primitive} and \textbf{society} to distinguish between the agents at the local and global levels, respectively, which we define in the context of their local and global environments and objectives:

\begin{definition}
A \textbf{primitive} $\primitive$ is a tuple $(\agentbidder, \agenttransformation_{\mathcal{T}})$ of a bidding policy $\agentbidder : \mathcal{S} \rightarrow \mathcal{B}$ and transformation $\agenttransformation_{\mathcal{T}}: \mathcal{S} \rightarrow \mathcal{S}$.
\end{definition}

\begin{definition}
A \textbf{society} $\Omega$ is a set of primitives $\primitive^{1:N}$.
\end{definition}

The \textbf{global environment} is an MDP that we call the \textbf{global MDP}, with state space $\mathcal{S}$ and discrete action space $\mathcal{A} = \{1, ..., N\}$ that indexes the primitives $\omega^{1:N}$.
The \textbf{local environment} is an auction that we call the \textbf{local auction} with auction item $s \in \mathcal{S}$ and bidding space $\mathcal{B} = [0, \infty)$.

The connection between the local and global environments is as follows.
Each state in the global MDP is an auction item for a different local auction.
The winning primitive $\winnerprimitive$ of the auction at state $s$ transforms $s$ into the next state $s'$ of the global MDP using its transformation $\phi_{\mathcal{T}}$, parameterized by the global MDP's transition function $\mathcal{T}$.
For each primitive $i$ at each state $s$, its \textbf{local objective} is the utility $\yourutility_s(\allbidders)$.
Its \textbf{local problem} is to maximize $\yourutility_s(\allbidders)$.
The \textbf{global objective} is the return $J(\pi_\Omega)$ in the global MDP of the \textbf{global policy} $\pi_\Omega$.
The \textbf{global problem} for the society is to maximize $J(\pi_\Omega)$.
We define the optimal \textbf{societal Q function} $Q^*_{\Omega}(s, \omega)$ as the expected return received from $\omega$ invoking its transformation $\phi_{\mathcal{T}}$ on $s$ and the society activating primitives optimally with respect to $J(\pi_\Omega)$ afterward.

Since all decisions made at the societal level are an emergent consequence of decisions made at the primitive level, the societal decision-making framework is a self-organization perspective on a broad range of sequential decision problems.
If each transformation $\phi_{\mathcal{T}}$ specifies a literal action, then societal decision-making is a decentralized re-framing of standard reinforcement learning (RL).
Societal decision-making also encompasses the decision problem of choosing $\phi_{\mathcal{T}}$s as options in semi-MDPs~\citep{sutton1999between} as well as choosing $\phi_{\mathcal{T}}$s as functions in a computation graph~\citep{chang2018automatically,rosenbaum2017routing,alet2018modular}.

We are interested in auction mechanisms and learning algorithms for optimizing the global objective as an emergent consequence of optimizing the local objectives.
Translating problems from one level of abstraction to another would provide a recipe for engineering a multi-agent system to achieve a desired global outcome and permit theoretical expectations on the nature of the equilibrium of the society, while giving us free choice on the architectures and learning algorithms of the primitive agents.
To this end, we next present an auction mechanism for which the dominant strategy equilibrium of the primitives coincides with the optimal policy of the society, which we develop into a class of decentralized RL algorithms in later sections.
\section{Mechanism Design for the Society} \label{sec:mechanism}
We first observe that to produce the optimal global policy, the optimal bidding strategy for each primitive at each local auction must be to bid their societal Q-value.
By defining each primitive's valuation of a state as its optimal societal Q-value at that state, we show that the Vickrey auction ensures the dominant strategy equilibrium profile of the primitives coincides with the optimal global policy.
Then we show that a market economy perspective on societal decision-making overcomes the need to assume knowledge of optimal Q-values, although weakens the dominant strategy equilibrium to a Nash equilibrium.
Lastly, we explain that adding redundant primitives to the society mitigates market bubbles by enforcing credit conservation.
Proofs are in the Appendix.

\subsection{Optimal Bidding}
We state what was observed informally in~\cite{baum1996toward}:
\begin{proposition} \label{proposition:optimal_policy}
Assume at each state $s$ the local auction allocates $\yourallocation(\mathbf{b})=1$ if $i$ wins and $\yourallocation(\mathbf{b})=0$ if $i$ loses.
Then all primitives $\yourprimitive$ bidding their optimal societal Q-values $\societalQ^*(s, \yourprimitive)$ collectively induce an optimal global policy.
\end{proposition}
This proposition makes the problem of self-organization concrete: getting the optimal behavior in the global MDP to emerge from the optimal behavior in the local auctions can be reduced to incentivizing the primitives to bid their optimal societal Q-value at every state.

\subsection{Dominant Strategies for Optimal Bidding} \label{sec:dominant_strategies}
To incentivize the primitives to bid optimally, we propose to define the primitives' valuations $\yourvalue_s$ for each state $s$ as their optimal societal Q-values $\societalQ^*(s, \yourprimitive)$ and use the Vickrey auction mechanism for each local auction.
\begin{theorem} \label{thm:dominant_strategy_equilibrium_Q}
If the valuations $\yourvalue_s$ for each state $s$ are the optimal societal Q-values $\societalQ^*(s, \yourprimitive)$, then the society's optimal global policy coincides with the primitives' unique dominant strategy equilibrium under the Vickrey mechanism.
\end{theorem}
Then, the utility $\yourutility_s(\allbidders)$ at each state $s$ that induces the optimal global policy, which we refer equivalently as $\winnerutility_s(\allprimitives)$ for the winning primitive $\winnerprimitive^i$ and $U_s^j(\allprimitives)$ for losing primitives $\omega^j$, is given by
\begin{align}
    \winnerutility_s(\allprimitives) = \societalQ^*(s, \winnerprimitive^i) - \max_{j\neq i} \mathbf{b}^j_s
\end{align}
and by $U_s^j(\allprimitives) = 0$ for losing primitives.

\subsection{Economic Transactions for Propagating $\societalQ^*$} \label{sec:economic_transactions}
We have so far defined optimal bidding with respect to societal decision-making and characterized the utilities as functions of $\societalQ^*$ for which such bidding is a dominant strategy.
We now propose to redefine the utilities without knowledge of $\societalQ^*$ by viewing the society as a market economy.

Monolithic frameworks for solving MDPs, such as directly optimizing the policy $J(\pi)$ with policy gradient methods, are analogous to \textbf{command-economies}, in which all production -- the transformation of past states $s_t$ into future states $s_{t+1}$ -- and wealth distribution -- the credit assignment of reward signals to parameters -- derive directly from single central authority -- the MDP objective.
In contrast, under the societal decision-making framework, the optimal global policy does \textit{not} derive directly from the MDP objective, but rather emerges implicitly as the equilibrium of the primitives optimizing their own local objectives.
We thus redefine the valuations $\yourvalue_s$ following the analogy of a \textbf{market economy}, in which production and wealth distribution are governed by the economic transactions between the primitives.

Specifically, we couple the local auctions at consecutive time-steps in the same game by defining the valuation $\yourvalue_{s_t}$ of primitive $\winnerprimitive^i$ for winning the auction item $s_t$ as the revenue it can receive in the auction at the next time-step by selling the product $s_{t+1}$ of executing its transformation $\phi_{\mathcal{T}}^i$ on $s_t$.
This compensation comes as the environment reward plus the (discounted) winning bid at the next time-step:
\begin{equation} \label{eqn:multistep_valuation}
\underset{\text{utility}}{\underbrace{\vphantom{\max_{j \neq i}}\winnerutility_{s_t}(\allprimitives)}} = \underset{\text{revenue, or valuation} \; \yourvalue_{s_t}}{\underbrace{\vphantom{\max_{j \neq i}}r(s_t, \winnerprimitive^i) + \gamma \cdot \max_k \mathbf{b}^k_{s_{t+1}}}} - \underset{\text{price}}{\underbrace{\max_{j \neq i} \mathbf{b}^j_{s_t}}}.
\end{equation} 
Analogous to a market economy, the revenue $\winnerprimitive^i$ receives for producing $s_{t+1}$ from $s_t$ depends on the price the winning primitive $\winnerprimitive^k$ at $t+1$ is willing to bid for $s_{t+1}$.
In turn, $\winnerprimitive^k$ sells $s_{t+2}$ to the winning primitive at $t+2$, and so on.
Ultimately currency is grounded in the reward.
Wealth is distributed based on what future primitives decide to bid for the fruits of the labor of information processing carried out by past primitives transforming one state to another.
\begin{definition}
A \textbf{Market MDP} is a global MDP in which all utilities are defined as in Equation~\ref{eqn:multistep_valuation}.
\end{definition}
As valuations now depend on the strategies of future primitives, the dominant strategy equilibrium from Theorem~\ref{thm:dominant_strategy_equilibrium_Q} must be weakened to a Nash equilibrium in the general case:
\begin{proposition} \label{proposition:nash_equilibrium}
In a Market MDP, it is a Nash equilibrium for every primitive to bid $\societalQ^*(s, \yourprimitive)$.
Moreover, if the Market MDP is finite horizon, then bidding $\societalQ^*(s, \yourprimitive)$ is the unique Nash equilibrium that survives iterated deletion of weakly dominated strategies.
\end{proposition}

\begin{figure}
    \centering
    \includegraphics[width=0.45\textwidth]{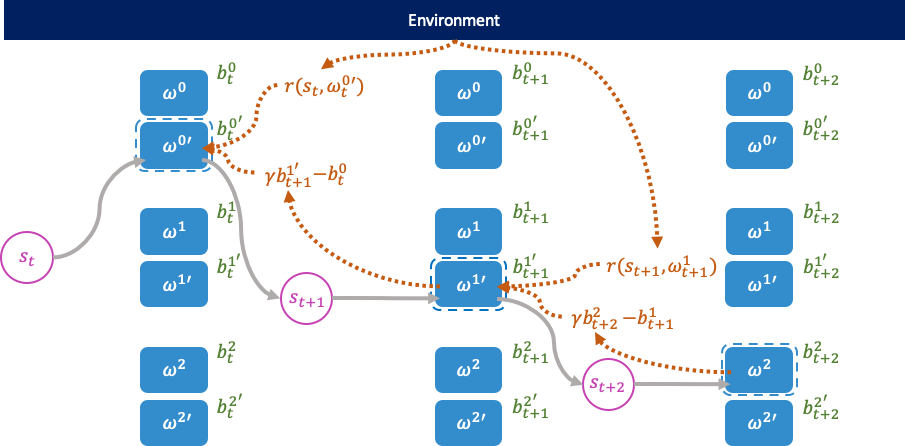}
    \caption{\small{\textbf{The cloned Vickrey society.}
    In this market economy of primitive agents, wealth is distributed not directly from the global MDP objective but based on what future primitives decide to bid for the fruits of the labor of information processing carried out by past primitives transforming one state to another.
    The primitive $\primitive_{t}^{0'}$ that wins the auction at time $t$ receives an environment reward $r(s_t, \primitive_{t}^{0'})$ as well as payment $b^{1'}_{t+1}$ from $\primitive_{t+1}^{1'}$ for transforming $s_t$ to $s_{t+1}$.
    By the Vickrey auction, the price $\primitive_{t+1}^{1'}$ pays to transform $s_{t+1}$ is the second highest bid $b_{t+1}^{1}$ at time $t+1$.
    Because each primitive $\omega^i$ and its clone $\omega^{i'}$ have the same valuations, their bids are equivalent and so credit is conserved through time.
    }}
    \label{fig:main}
\end{figure}

\subsection{Redundancy for Credit Conservation} \label{sec:redundancy_for_credit_conservation}
In general, credit is not conserved in the Market MDP: the winning primitive at time $t-1$ gets paid an amount equal to the highest bid at time $t$, but the winner at time $t$ only pays an amount equal to the second highest bid at time $t$.
At time $t$, $\omega^i$ could bid arbitrarily higher than $\max_{j \neq i} \mathbf{b}^j$ without penalty, which distorts the valuations for primitives that bid before time $t$, creating a market bubble.

To prevent market bubbles, we propose a modification to the society, which we will call a \textbf{cloned society}, for enforcing credit conservation -- for all transformations $\phi_{\mathcal{T}}$ initialize at least two primitives that share the same $\phi_{\mathcal{T}}$:
\begin{lemma} \label{lemma:credit_conservation}
For a cloned society, at the Nash equilibrium specified in Proposition~\ref{proposition:nash_equilibrium}, what the winning primitive $\winnerprimitive^i$ at time $t$ receives from the winning primitive $\winnerprimitive^k$ at $t+1$ is exactly what $\winnerprimitive^k$ pays: $\mathbf{b}^k_{s_{t+1}}$.
\end{lemma}
We now state our main result.
\begin{theorem} \label{theorem:credit_optimal_equilibrium}
Define a \textbf{cloned Vickrey society} as a cloned society that solves a Market MDP.
Then it is a Nash equilibrium for every primitive in the cloned Vickrey society to bid $\societalQ^*(s, \yourprimitive)$.
In addition, the price that the winning primitive pays for winning is equivalent to what it bid.
\end{theorem}
The significance of Theorem~\ref{theorem:credit_optimal_equilibrium} is that guaranteeing truthful bidding of societal Q-values decouples the analysis of the local problem within each time-step from that of the global problem across time-steps, which opens the possibility for designing learners to reach a Nash equilibrium that we know exists.
Without such a separation of these two levels of abstraction the entire society must be analyzed as a repeated game through time -- a non-trivial challenge.

\section{From Equilibria to Learning Objectives} \label{sec:learning}
So far the discussion has centered around the quality of the equilibria, which assumes the primitives know their own valuations.
We now propose a class of decentralized RL algorithms for learning optimal bidding behavior without assuming knowledge of the primitives' valuations.

Instead of learning to optimize the MDP objective directly, we propose to train the primitives' parameterized bidding policies to optimize their utilities in Equation~\ref{eqn:multistep_valuation}, yielding a class of decentralized RL algorithms for optimizing the global RL objective that is agnostic to the choice of RL algorithm used to train each primitive.
By Theorem~\ref{theorem:credit_optimal_equilibrium}, truthful bidding for all agents is one global optimum to all agent's local learning problems that also serves as a global optimum for the society as whole.
In the special case where the transformation $\phi_{\mathcal{T}}$ is a literal action, this class of decentralized RL algorithms can serve as an alternative to any standard algorithm for solving discrete-action MDPs.
An on-policy learning algorithm is presented in Appendix~\ref{appdx:decentralized_rl}.

\subsection{Local Credit Assignment in Space and Time}
The global problem requires a solution that is global in space, because the society must collectively work together, and global in time, because the society must maximize expected return across time-steps.
But an interesting property of using the auction utility in Equation~\ref{eqn:multistep_valuation} as an RL objective is that given redundant primitives it takes the form of the Bellman equation, thereby implicitly coupling all primitives together in space and time.
Thus each primitive need only optimize for its immediate utility at a each time-step without needing to optimize for its own future utilities.
This class of decentralized RL algorithms thus implicitly finds a global solution in space and time using only credit assignment that is \emph{local in space}, because transactions are only between individual agents, and \emph{local in time}, because each primitive need only solve a contextual bandit problem at each time-step.

\subsection{Redundancy for Avoiding Suboptimal Equilibria}
A benefit of casting local auction utilities as RL objectives is the practical development of learning algorithms that need not assume oracle knowledge of valuations, but unless care is taken with each primitives' learning environments, the society may not converge to the globally optimal Nash equilibrium described in Section~\ref{sec:mechanism}.
As an example of a suboptimal equilibrium, in a Market MDP with two primitives, even if $\mathbf{v}^1=1, \mathbf{v}^2=2$, it is a Nash equilibrium for $\mathbf{b}^1=100, \mathbf{b}^2=0$.
Without sufficient competitive pressure to not overbid or underbid, a rogue winner lacks the risk of losing to other primitives with similarly close valuations.
Fortunately, the redundancy of a cloned Vickrey society serves a dual purpose of not only preventing market bubbles but also introducing competitive pressure in the primitives' learning environments in the form of other clones.
\section{Experiments} \label{sec:experiments}
Now we study how well the cloned Vickrey society can recover the optimal societal Q-function as its Nash equilibrium.
We compare several implementations of the cloned Vickrey society against baselines across simple tabular environments in Section~\ref{sec:numerical_simulations}, where the transformations $\phi_\mathcal{T}$ are literal actions.
Then in Section~\ref{sec:semi_mdps_comp_graphs} we demonstrate the broad applicability of the cloned Vickrey society for learning to select options in semi-MDPs and composing functions in a computation graph.
Moreover, we show evidence for the advantages of the societal decision making framework over its monolithic counterpart in transferring to new tasks.

Although our characterization of the equilibrium and the learning algorithm are agnostic to the implementation of the primitive, in our experiments the bidding policy $\psi$ is implemented as a neural network that maps the state to the parameters of a Beta distribution, from which a bid is sampled. 
We used proximal policy optimization (PPO)~\citep{schulman2017proximal} to optimize the bidding policy parameters.

\subsection{Numerical Simulations} \label{sec:numerical_simulations}
In the following simulations we ask: (1) How closely do the bids the primitives learn match their optimal societal Q-values? (2) Does the solution to the global objective emerge from the competition among the primitives? (3) How does redundancy affect the solutions the primitives converge to?

We first consider the \textit{Market Bandit}, the simplest global MDP with one state, and compare the cloned Vickrey society with societies with different auction mechanisms.
Next we consider the \textit{Chain} environment, a sparse-reward multi-step global MDP designed to study market bubbles.
Last we consider the \textit{Duality} environment, a multi-step global MDP designed to study suboptimal equilibria.
We use \textit{solitary society} to refer to a society without redundant clones.

\subsubsection{Market Bandits and Auction Mechanisms} \label{sec:empirical_bandits}
\begin{figure}[t]
\begin{minipage}[t]{0.23\textwidth}
\vspace{0pt}
\centering
\includegraphics[width=1.04\textwidth,center]{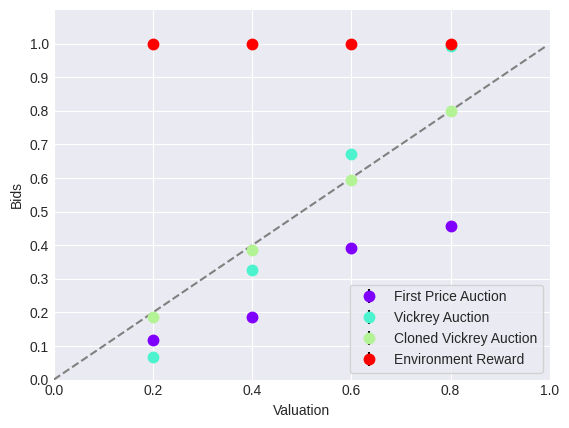}
\subcaption{}
\label{fig:bandit_correlation}
\end{minipage}
\begin{minipage}[t]{0.23\textwidth}
\vspace{0pt}
\centering
\includegraphics[width=1.04\textwidth,center]{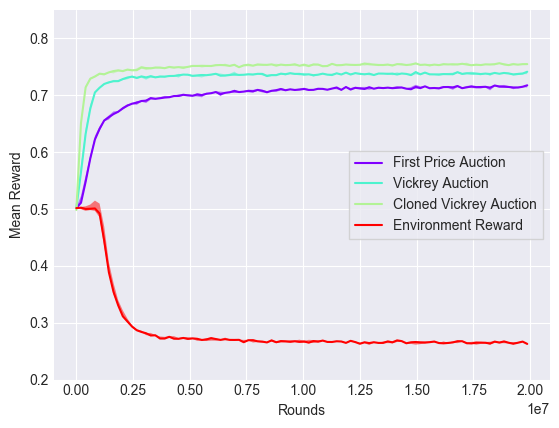}
\subcaption{}
\label{fig:bandit_learning_curve}
\end{minipage}
\caption{\small{
\textbf{Market Bandits.} 
We compare the cloned Vickrey society (Cloned Vickrey Auction) against solitary societies that use the first-price auction mechanism (first price auction), the Vickrey auction mechanism (Vickrey Auction), and a mechanism whose utility is only the environment reward (Environment Reward).
The dashed line in (a) indicates truthful bidding of valuations.
The cloned Vickrey society's bids are closest to the true valuations which also translates into the best global policy (b).
}}
\label{fig:bandit_results}
\end{figure}

The simulation primarily studies question (1) by comparing the auction mechanism of the cloned Vickrey society against other mechanisms in eliciting bids that match optimal societal Q-values.
We compare against \textit{first price auction}, based on~\citet{holland1985properties}, a solitary society that uses the first price auction mechanism for the local auction, in which the winning primitive pays a price of their bid, rather the second highest bid;
against \textit{Vickrey Auction}, a solitary version of the cloned Vickrey society;
and against \textit{Environment Reward}, a baseline solitary society whose utility function uses only the environment reward, with no price term $P^i(\mathbf{b})$.

The environment is a four-armed Market Bandit whose arms correspond to transformations $\phi$ that deterministically yield reward values of $0.2$, $0.4$, $0.6$, and $0.8$.
Solitary societies thus would have four primitives, while cloned societies eight.
Since we are mainly concerned with question (1), we stochastically drop out a subset of primitives at every round to give each primitive a chance to win and learn its value.

In the Market Bandit, the arm rewards directly specify the primitives' valuations, so if the primitives successfully learned their valuations, we would expect them to learn to bid values exactly at $0.2$, $0.4$, $0.6$, and $0.8$.
Figure~\ref{fig:bandit_correlation} shows that the primitives in the cloned Vickrey society learn to bid most closely to their true valuations, which also translates into the best global policy in Figure~\ref{fig:bandit_learning_curve}, answering question (2).
To address question (3), we observe that the solitary Vickrey society's bids are more spread out than those of the cloned Vickrey society because there is no competitive pressure to learn the valuations exactly.
This is not detrimental when the global MDP has only one time-step, but the next section shows that the lack of redundancy creates market bubbles that result in a suboptimal global policy.

\begin{figure}
\begin{minipage}[t]{0.18\textwidth}
\vspace{0pt}
\centering
\includegraphics[width=0.9\textwidth,center]{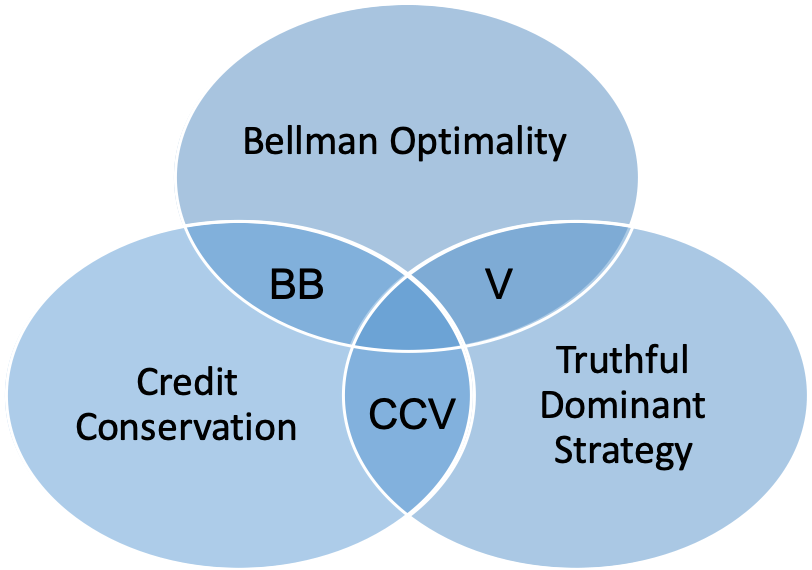}
\end{minipage}
\begin{minipage}[t]{0.25\textwidth}
\vspace{0pt}
\centering
\begin{tabular}{ c|c|c } 
\hline
 & $\winnerprimitive_t$ Receives & $\winnerprimitive_{t+1}$ Pays \\ 
\hline
 \emph{CCV} & $\mathbf{b}'_{t+1}$ & $\mathbf{b}'_{t+1}$ \\ 
 \bucketbrigade & $\hat{\mathbf{b}}_{t+1}$ & $\hat{\mathbf{b}}_{t+1}$ \\ 
 \emph{V} & $\hat{\mathbf{b}}_{t+1}$ & $\mathbf{b}'_{t+1}$ \\ 
\hline
\end{tabular}
\end{minipage}
\caption{\small{
\textbf{Implementations.}
The table shows the bid price that temporally consecutive winners $\winnerprimitive_{t+1}$ and $\winnerprimitive_t$ pay and receive based on three possible implementations of the cloned Vickrey society: \textit{CCV}, \textit{BB}, \textit{V}, with tradeoffs depicted in the Venn diagram.
We use $\hat{\mathbf{b}}_{t+1}$ and $\mathbf{b}_{t+1}'$ to denote the highest and second highest bids at time $t+1$ respectively.
}}
\label{fig:possible_implementations}
\end{figure}

\begin{figure}[h]
\begin{subfigure}{.15\textwidth}
  \centering
    \includegraphics[width=\linewidth]{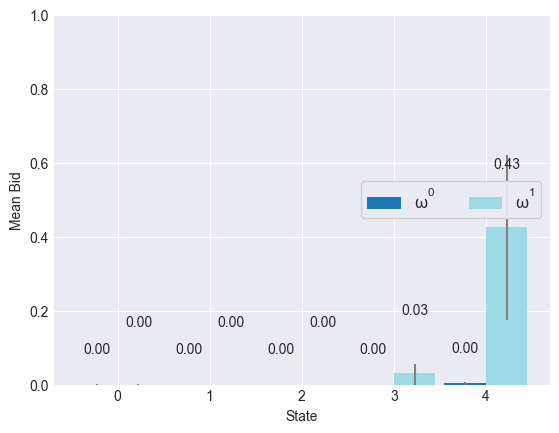}
  \caption{CCV solitary}
  \label{fig:ccv_1}
\end{subfigure}
\hfill
\begin{subfigure}{.15\textwidth}
  \centering
    \includegraphics[width=\linewidth]{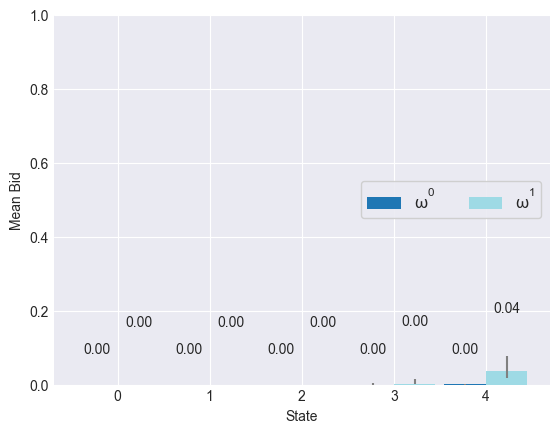}
  \caption{BB solitary}
  \label{fig:bb_1}
\end{subfigure}
\hfill
\begin{subfigure}{.15\textwidth}
  \centering
    \includegraphics[width=\linewidth]{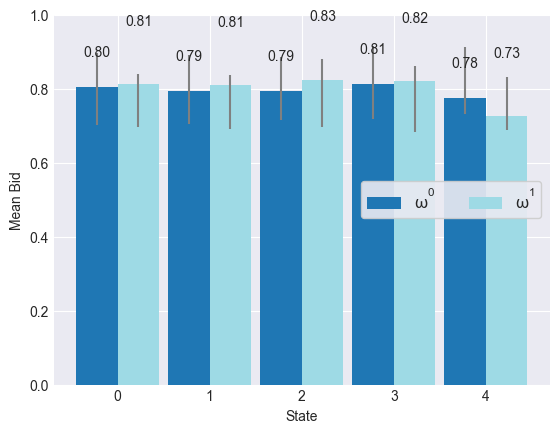}
  \caption{V solitary}
  \label{fig:v_1}
\end{subfigure}
\hfill
\begin{subfigure}{.15\textwidth}
  \centering
    \includegraphics[width=\linewidth]{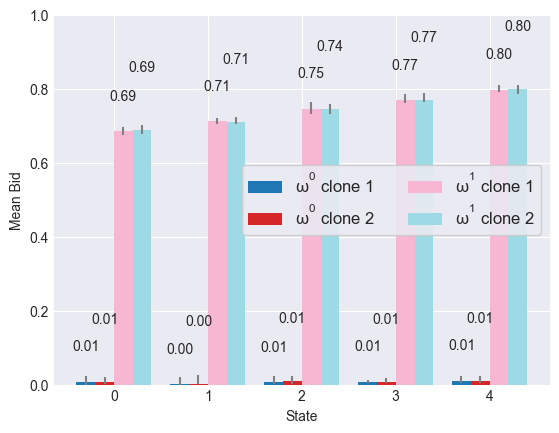}
  \caption{CCV clone}
  \label{fig:ccv_2}
\end{subfigure}
\hfill
\begin{subfigure}{.15\textwidth}
  \centering
  \includegraphics[width=\linewidth]{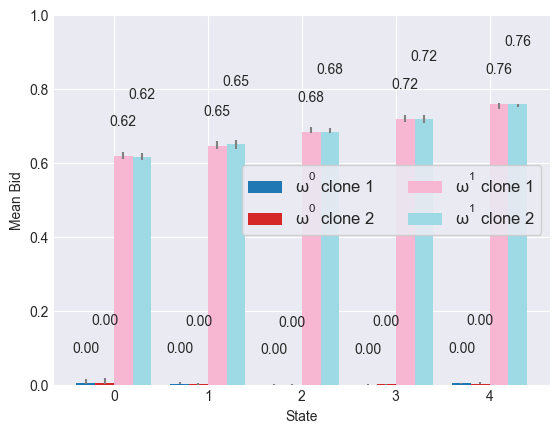}
  \caption{BB clone}
  \label{fig:bb_2}
\end{subfigure}
\hfill
\begin{subfigure}{.15\textwidth}
  \centering
  \includegraphics[width=\linewidth]{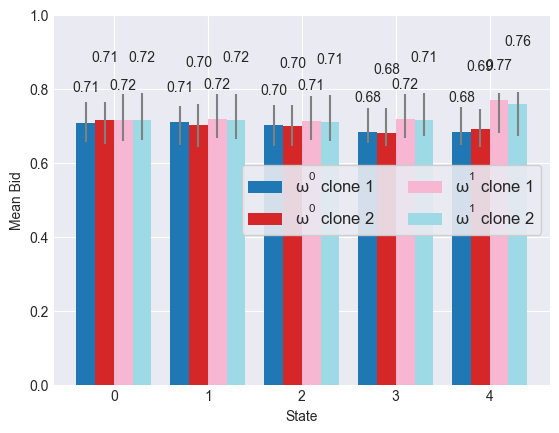}
  \caption{V clone}
  \label{fig:v_2}
\end{subfigure}
\caption{\small{
\textbf{Learned Bidding Strategies for \textit{Chain}.}
We organize the analysis by distinguishing between the credit-conserving (\textit{CCV} and \textit{BB}) and the non-credit-conserving (\textit{V}) implementations.
The solitary \textit{CCV} (a) and \textit{BB} (b) implementations learn to bid very close to $0$: \textit{CCV} because the valuation for a primitive at $t$ is only the second-highest bid at $t+1$, resulting in a rapid decay in the valuations leftwards down the chain; \textit{BB} because each primitive is incentivized to pay as low of a price for winning as possible.
The cloned \textit{CCV} (d) and \textit{BB} (e) implementations learn to implement a form of return decomposition~\citep{arjona2019rudder} that redistributes the terminal reward into a series of positive payoffs back through the chain, each agent getting paid for contributing to moving the society closer to the goal state, where the \textit{CCV} implementation's bids are closer to the optimal societal Q-value than those of the \textit{BB} implementation.
Because both the solitary (c) and cloned (f) versions of the \textit{V} implementations do not conserve credit, they learn to bid close to the optimal societal Q-value, but both suffer from market bubbles where the primitive for going left bids higher than the primitive for going right, even though the optimal global policy is to keep moving right.
}}
\label{fig:chain_bids}
\end{figure}

\subsubsection{Market MDPs} \label{sec:empirical_mdps}
Now we consider MDPs that involve multiple time-steps to understand how learning is affected when primitives' valuations are defined by the bids of future primitives.
Redundancy theoretically makes what the winner $\winnerprimitive_{t}$ receives and $\winnerprimitive_{t+1}$ equivalent, but the stochasticity of the bid distribution yields various possible implementations for the cloned Vickrey society in the market MDP-- \textit{bucket brigade} (\emph{BB}), \textit{Vickrey} (\emph{V}), and \textit{credit conserving Vickrey} (\emph{CCV}), summarized in Figure~\ref{fig:possible_implementations}, with different pros and cons.
Note that the solitary bucket brigade society is a multiple time-step analog of the \textit{first price auction} in Section~\ref{sec:empirical_bandits}.
We aim to empirically test which of the implementations of the cloned Vickrey society yields best local and global performance.

\begin{figure*}
    \centering
    \includegraphics[width=\textwidth,center]{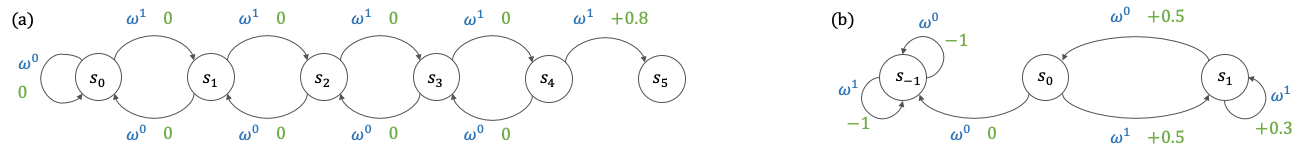}
    \caption{
    \small{
    \textbf{Multi-Step MDPs.}
    (a) In the \emph{Chain} environment, the society starts at state $s_0$ and the goal state is $s_5$.
    Only activating primitive $\omega_1$ at state $s_4$ yields reward.
    The optimal global policy is to directly move right by continually activating $\omega_1$.
    Without credit conservation, the society may get stuck going back and forth between $s_0$ and $s_4$ without reaching the goal.
    (b) In the \emph{Duality} environment, the society starts at state $s_0$. 
    $s_{-1}$ is an absorbing state with perpetual negative rewards.
    The optimal societal policy is to cycle between $s_0$ and $s_1$ to receive unbounded reward, but without redundant primitives, the society may end up in a suboptimal perpetual self-loop at $s_1$. 
    }
    }
    \label{fig:multistep}
\end{figure*}

\begin{figure} 
\begin{minipage}[t]{0.23\textwidth}
\vspace{0pt}
\centering
\includegraphics[width=0.9\textwidth,center]{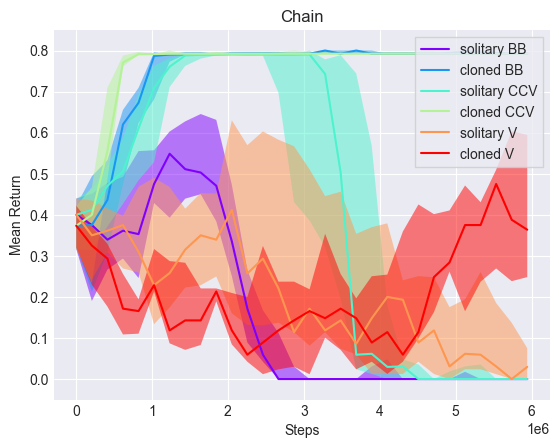}
\subcaption{Chain}
\label{fig:chain_curve}
\end{minipage}
\hfill
\begin{minipage}[t]{0.23\textwidth}
\vspace{0pt}
\centering
\includegraphics[width=0.9\textwidth,center]{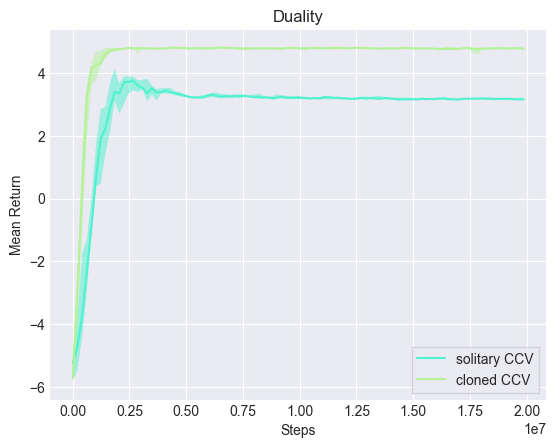}
\subcaption{Duality}
\label{fig:Duality_curve}
\end{minipage}
\caption{\small{
\textbf{Multi-Step MDP Global Learning Curves.}
We observe that cloned societies are more robust against suboptimal equilibria than solitary societies.
Furthermore the cloned \textit{CCV} implementation achieves the best sample efficiency, suggesting that truthful bidding and credit-conservation are important properties to enforce for enabling the optimal global policy to emerge.
}}
\label{fig:mdp_results}
\end{figure}

\paragraph{The \textit{Chain} environment and market bubbles.}
A primary purpose of the \textit{Chain} environment (Figure~\ref{fig:multistep}a) is to study the effect of redundancy on mitigating market bubbles in the bidding behavior and how that affects global optimality.
The \textit{Chain} environment is a sparse reward finite-horizon environment which initializes the society at $s_0$ on the left, yields a terminal reward of $0.8$ if the society enters the goal state $s_5$ on the right, and ends the episode after 20 steps if the society does not reach the goal.
\textit{Chain} thus tests the ability of the society to propagate utility from future agents to past agents in the absence of an immediate environment reward signal.
We compare the bidding behaviors of the \textit{BB}, \textit{V}, and \textit{CCV} implementations of the cloned Vickrey society, as well as those of their solitary counterparts, in Figure~\ref{fig:multistep} and the society's global learning curve in Figure~\ref{fig:chain_curve}.

\begin{figure}[ht]
\begin{subfigure}{.15\textwidth}
  \centering
  \includegraphics[width=\linewidth]{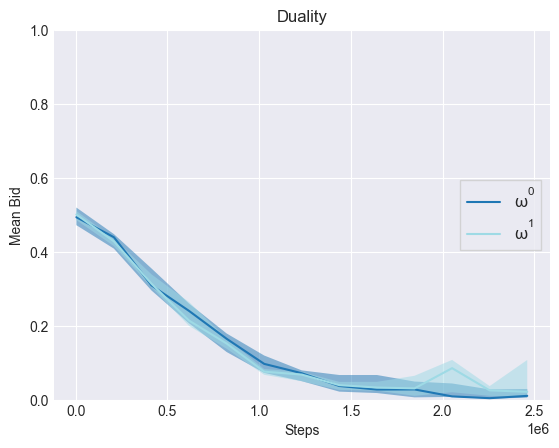}
  \caption{solitary: $s_{-1}$}
  \label{fig:1_s-1}
\end{subfigure}
\hfill
\begin{subfigure}{.15\textwidth}
  \centering
  \includegraphics[width=\linewidth]{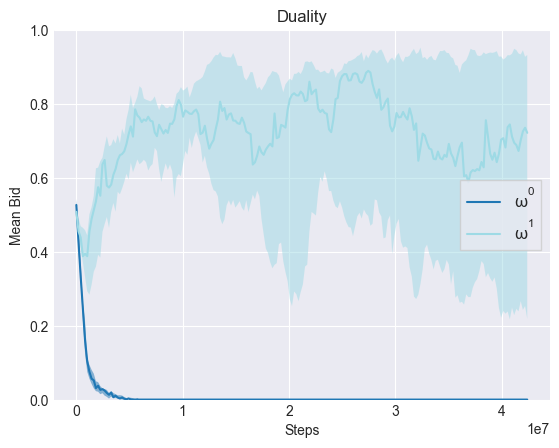}
  \caption{solitary: $s_{0}$}
  \label{fig:1_s0}
\end{subfigure}
\hfill
\begin{subfigure}{.15\textwidth}
  \centering
  \includegraphics[width=\linewidth]{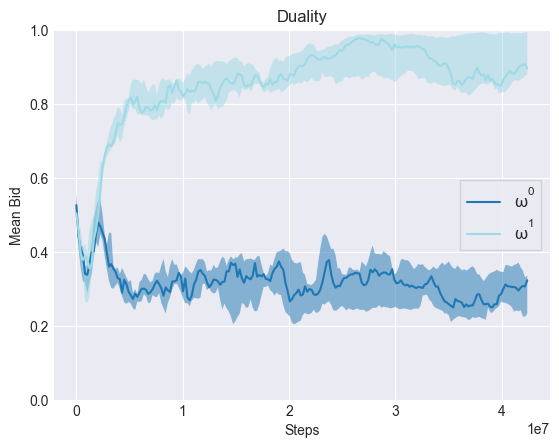}
  \caption{solitary: $s_{1}$}
  \label{fig:1_s1}
\end{subfigure}
\hfill
\begin{subfigure}{.15\textwidth}
  \centering
  \includegraphics[width=\linewidth]{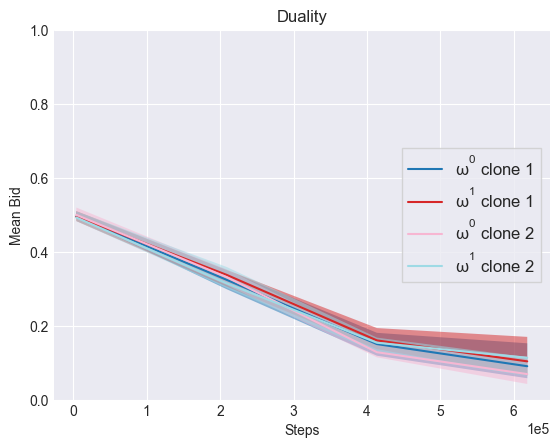}
  \caption{cloned: $s_{-1}$}
  \label{fig:2_s-1}
\end{subfigure}
\hfill
\begin{subfigure}{.15\textwidth}
  \centering
  \includegraphics[width=\linewidth]{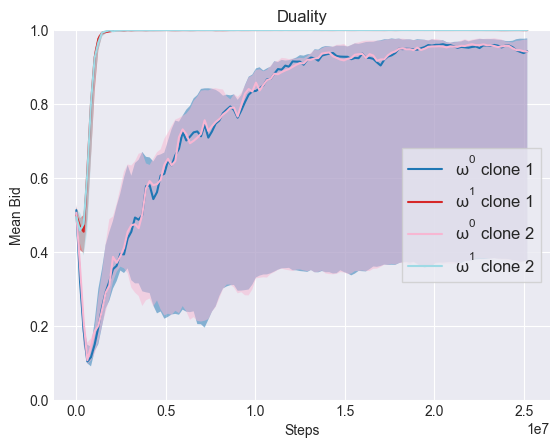}
  \caption{cloned: $s_{0}$}
  \label{fig:2_s0}
\end{subfigure}
\hfill
\begin{subfigure}{.15\textwidth}
  \centering
  \includegraphics[width=\linewidth]{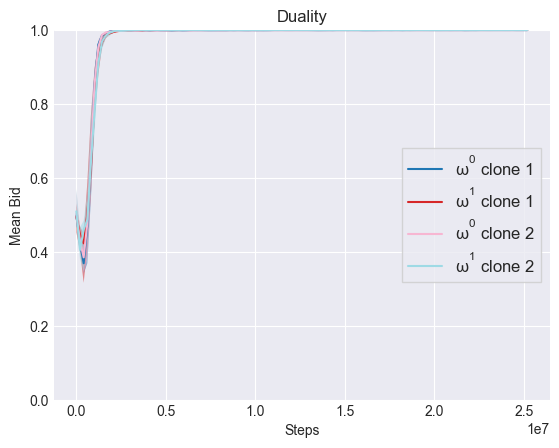}
  \caption{cloned: $s_{1}$}
  \label{fig:2_s1}
\end{subfigure}
\caption{\small{
\textbf{CCV Bidding Curves for \textit{Duality}.}
Each column shows the bidding curves of the solitary (top row) and cloned (bottom row) \textit{CCV} societies for states $s_{-1}$, $s_0$, and $s_1$.
Without redundant primitives to force the second-highest and highest valuations to be equal, the dominant strategy of truthful bidding may not coincide with the globally optimal policy because the solitary \textit{CCV} implementation does not guarantee Bellman optimality.
The bidding curves in (c) show that $\primitive^1$ learns a best response of bidding \emph{higher} than primitive $\primitive^0$ at state $s_1$, even though it would be globally optimal for the society if $\primitive^0$ wins at $s_1$.
Adding redundant primitives causes the second-highest and highest valuations to be equal, causing $\primitive^0$ to learn to bid highly as well at $s_1$, which results in a more optimal return as shown in Figure~\ref{fig:Duality_curve}.
}}
\label{fig:duality_bids}
\end{figure}

To answer question (1), we observe that redundancy indeed prevents market bubbles, with the cloned \textit{CCV} implementation bidding closet to the optimal societal Q values.
Details are in the caption of Figure~\ref{fig:chain_bids}.
When we consider the society's global learning curve in Figure~\ref{fig:chain_curve}, the answers to questions (2) and (3) go hand-in-hand: the solitary societies fail to find the globally optimal policy and the cloned \textit{CCV} implementation has the highest sample efficiency.

\paragraph{The \textit{Duality} environment and suboptimal equilibria.}
The \textit{Chain} environment experiments suggested a connection between lack of redundancy and globally suboptimal equilibria, the subtleties of which we explore further in the \textit{Duality} environment (Figure~\ref{fig:multistep}b).
The \textit{CCV} implementation has yielded the best performance so far but does not guarantee Bellman optimality (Figure~\ref{fig:possible_implementations}) without redundant primitives.
We show that in the \textit{Duality} environment, without redundant primitives the dominant strategy equilibrium would lead the society to get stuck in a self-loop at $s_1$ indefinitely, even though the global optimal solution would be to cycle back and forth between $s_0$ and $s_1$.
The reasoning for this is explained in Appendix~\ref{appdx:duality_details}.
Figure~\ref{fig:Duality_curve} indeed shows that a society with redundant primitives learns a better equilibrium than without.

\begin{figure*}
    \centering
    \includegraphics[width=0.8\textwidth,center]{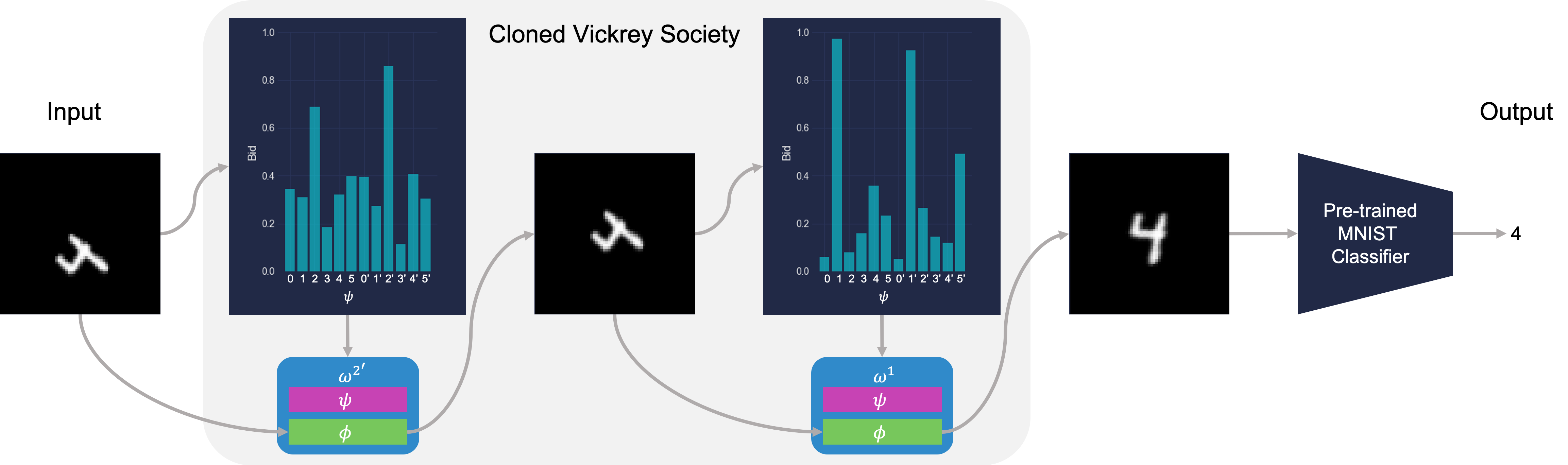}
    \caption{
    \small{
    \textbf{Mental Rotation.}
    The cloned Vickrey society learns to transform the image into a form that can be classified correctly with a pre-trained classifier by composing two of six possible affine transformations of rotation and translation. Clones are indicated by an apostrophe. 
    In this example, the society activated primitive $\omega^{2'}$ to translate the digit up then primitive $\omega^{1}$ to rotate the digit clockwise.
    Though the bidding policies $\psi^i$ and ${\psi^i}'$ of the clones $\omega^i$ and ${\omega^i}'$ have the same parameters, their sampled bids may be different because the bidding policies are stochastic.
    }
    }
    \label{fig:mnist}
\end{figure*}

\subsection{Semi-MDPs and Computation Graphs} \label{sec:semi_mdps_comp_graphs}
One benefit of framing global decision-making from the perspective of local economic transactions is that the same societal decision-making framework and learning algorithms can be used regardless of the type of transformation $\phi_{\mathcal{T}}$.
We now show in the \textit{Two Rooms} environment that the cloned Vickrey society can learn more efficiently than a monolithic counterpart to select among pre-trained options to solve a \texttt{gym-minigrid}~\citep{gym_minigrid} navigation task that involves two rooms separated by a bottleneck state.
We also show in the \textit{Mental Rotation} environment that the cloned Vickrey society can learn to dynamically compose computation graphs of pre-specified affine transformations for classifying spatially transformed MNIST digits.
We use the cloned \textit{CCV} society for these experiments.

\subsubsection{Transferring with Options in Semi-MDPs} \label{sec:empirical_hrl}
\begin{figure}
    \centering
    \includegraphics[width=0.45\textwidth]{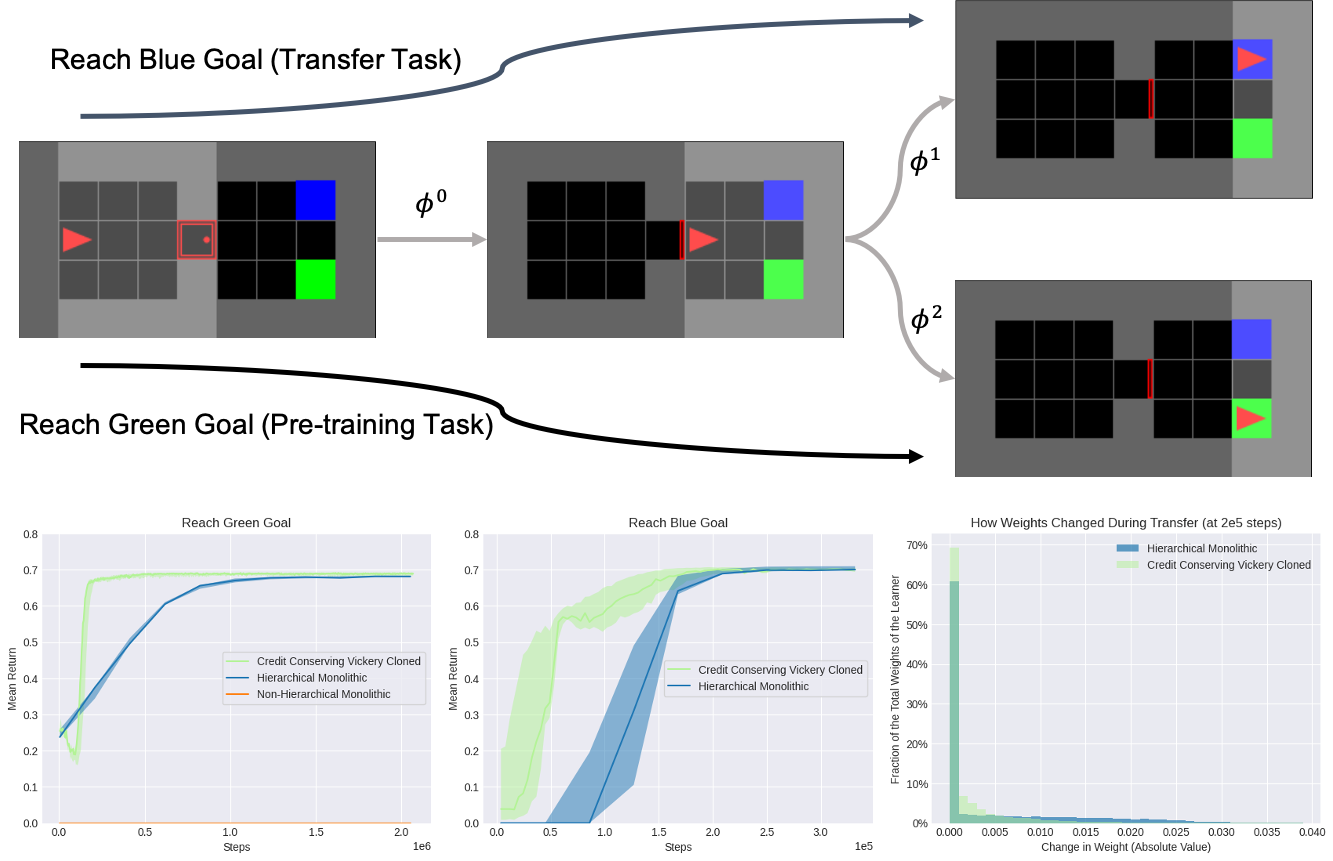}
    \caption{\small{
    \textbf{Two Rooms.}
    The cloned Vickrey society adapts more quickly than the hierarchical monolithic baseline in both the pre-training and the transfer tasks.
    The bottom-right figure, which is a histogram of the absolute values of how much the weights have shifted from fine-tuning on the transfer task, shows that more weights shift, and to a larger degree, in the hierarchical monolithic baseline than in our method.
    This seems to suggest that the cloned Vickrey society is a more modular learner than the hierarchical monolithic baseline.
    The non-hierarchical monolithic baseline does not learn to solve the task from scratch.
    }}
    \label{fig:hrl}
\end{figure}

We construct a two room environment, \textit{Two Rooms}, which requires opening the red door and reaching either a green goal or blue goal.
The transformations $\phi_{\mathcal{T}}$ are subpolicies that have been pre-trained to open the red door, reach the green goal, and reach the blue goal.
In the pre-training task, only reaching the green goal gives a non-zero terminal reward and reaching the blue goal does not give reward.
In the transfer task, the rewards for reaching the green and blue goals are switched.
We compared the cloned Vickrey society against a non-hierarchical monolithic baseline that selects among the low-level \texttt{gym-minigrid} actions as well as a hierarchical monolithic baseline that selects among the pre-trained subpolicies.
Both baseline policies sample from a Categorical distribution and are trained with PPO.
The cloned Vickrey society is more sample efficient in learning on the pre-training task and significantly faster in adapting to the transfer task (Figure~\ref{fig:hrl}).

Our hypothesized explanation for this efficiency is that the local credit assignment mechanisms of a society parallelize learning across space and time, a property that is not true of the global credit assignment mechanisms of a monolithic learner.
To test this hypothesis, we observe in the bottom-right of Figure~\ref{fig:hrl} that not only has a higher percentage of the hierarchical monolithic baseline's weights shifted during transfer compared to our method but they also shift to a larger degree, which suggests that the hierarchical monolithic baseline's weights are more globally coupled and perhaps thereby slower to transfer.
While this analysis is not comprehensive, it is a suggestive result that motivates future work for further study.

\subsubsection{Composing Dynamic Computation Graphs} \label{sec:empirical_mnist}
We adapt the Image Transformations task from ~\citet{chang2018automatically} as what we call the \textit{Mental Rotation} environment (Figure~\ref{fig:mnist}), in which MNIST images have been transformed with a composition of one of two rotations and one of four translations.
There are $60,000 \times 8 = 240,000$ possible unique inputs, meaning learning an optimal global policy involves training primitives across $240,000$ local auctions.
The transformations $\phi_{\mathcal{T}}$ are pre-specified affine transformations.
The society must transform the image into a form that can be classified correctly with a pre-trained classifier, with a terminal reward of 1 if the predicted label is correct and 0 otherwise.
The cloned Vickrey society converges to a mean return of 0.933 with a standard deviation of 0.014.
\section{Conclusion}
This work formally defines the societal decision-making framework and proves the optimality of a society -- the cloned Vickrey society -- for solving a problem that was first posed in the AI literature in the 1980s: to specify the incentive structure that causes the global solution of a society to emerge as the equilibrium strategy profile of self-interested agents.
For training the society, we further proposed a class of decentralized reinforcement learning algorithms whose global objective decouples in space and time into simpler local objectives.
We have demonstrated the generality of this framework for selecting actions, options, and computations as well as its potential advantages for transfer learning.

The generality of the societal decision-making framework opens much opportunity for future work in decentralized reinforcement learning.
A society's inherent modular structure suggests the potential of reformulating problems with global credit-assignment paths into problems with much more local credit-assignment paths that offer more parallelism and independence in the training of different components of a learner.
Understanding the learning dynamics of multi-agent societies continues to be an open problem of research.
It would be exciting to explore algorithms for constructing and learning societies in which the primitives are also societies themselves.
We hope that the societal decision-making framework and its associated decentralized reinforcement learning algorithms provide a foundation for future work exploring the potential of creating AI systems that reflect the collective intelligence of multi-agent societies.
\section*{Acknowledgements}
The authors are grateful to the anonymous reviewers, Michael Janner, and Glen Berseth for their feedback on this paper, as well as Aviral Kumar, Sam Toyer, Anirudh Goyal, Marco Cusumano-Towner, Abhishek Gupta, and Dinesh Jayaraman for helpful discussions.
This research was supported by AFOSR grant FA9550-18-1-0077, NSF CCF-1717899, and Google Cloud Platform.
MC is supported by the National Science Foundation Graduate Research Fellowship Program.
SK is supported by Berkeley Engineering Student Services.

\bibliography{bib}

\begin{thebibliography}{33}
\providecommand{\natexlab}[1]{#1}
\providecommand{\url}[1]{\texttt{#1}}
\expandafter\ifx\csname urlstyle\endcsname\relax
  \providecommand{\doi}[1]{doi: #1}\else
  \providecommand{\doi}{doi: \begingroup \urlstyle{rm}\Url}\fi

\bibitem[Alet et~al.(2018)Alet, Lozano-P{\'e}rez, and
  Kaelbling]{alet2018modular}
Alet, F., Lozano-P{\'e}rez, T., and Kaelbling, L.~P.
\newblock Modular meta-learning.
\newblock \emph{arXiv preprint arXiv:1806.10166}, 2018.

\bibitem[Arjona-Medina et~al.(2019)Arjona-Medina, Gillhofer, Widrich,
  Unterthiner, Brandstetter, and Hochreiter]{arjona2019rudder}
Arjona-Medina, J.~A., Gillhofer, M., Widrich, M., Unterthiner, T.,
  Brandstetter, J., and Hochreiter, S.
\newblock Rudder: Return decomposition for delayed rewards.
\newblock In \emph{Advances in Neural Information Processing Systems}, pp.\
  13544--13555, 2019.

\bibitem[Balduzzi(2014)]{balduzzi2014cortical}
Balduzzi, D.
\newblock Cortical prediction markets.
\newblock \emph{arXiv preprint arXiv:1401.1465}, 2014.

\bibitem[Baum(1996)]{baum1996toward}
Baum, E.~B.
\newblock Toward a model of mind as a laissez-faire economy of idiots.
\newblock In \emph{ICML}, pp.\  28--36, 1996.

\bibitem[Bellmann(1957)]{bellmann1957dynamic}
Bellmann, R.
\newblock Dynamic programming princeton university press.
\newblock \emph{Princeton, NJ}, 1957.

\bibitem[Braitenberg(1986)]{braitenberg1986vehicles}
Braitenberg, V.
\newblock \emph{Vehicles: Experiments in synthetic psychology}.
\newblock 1986.

\bibitem[Brooks(1991)]{brooks1991intelligence}
Brooks, R.~A.
\newblock Intelligence without representation.
\newblock \emph{Artificial intelligence}, 47\penalty0 (1-3):\penalty0 139--159,
  1991.

\bibitem[Chang et~al.(2018)Chang, Gupta, Levine, and
  Griffiths]{chang2018automatically}
Chang, M.~B., Gupta, A., Levine, S., and Griffiths, T.~L.
\newblock Automatically composing representation transformations as a means for
  generalization.
\newblock \emph{arXiv preprint arXiv:1807.04640}, 2018.

\bibitem[Chen et~al.(2009)Chen, Deng, and Teng]{chen2009settling}
Chen, X., Deng, X., and Teng, S.-H.
\newblock Settling the complexity of computing two-player nash equilibria.
\newblock \emph{Journal of the ACM (JACM)}, 56\penalty0 (3):\penalty0 1--57,
  2009.

\bibitem[Chevalier-Boisvert et~al.(2018)Chevalier-Boisvert, Willems, and
  Pal]{gym_minigrid}
Chevalier-Boisvert, M., Willems, L., and Pal, S.
\newblock Minimalistic gridworld environment for openai gym.
\newblock \url{https://github.com/maximecb/gym-minigrid}, 2018.

\bibitem[Daskalakis et~al.(2009)Daskalakis, Goldberg, and
  Papadimitriou]{daskalakis2009complexity}
Daskalakis, C., Goldberg, P.~W., and Papadimitriou, C.~H.
\newblock The complexity of computing a nash equilibrium.
\newblock \emph{SIAM Journal on Computing}, 39\penalty0 (1):\penalty0 195--259,
  2009.

\bibitem[Goyal et~al.(2019)Goyal, Sodhani, Binas, Peng, Levine, and
  Bengio]{goyal2019reinforcement}
Goyal, A., Sodhani, S., Binas, J., Peng, X.~B., Levine, S., and Bengio, Y.
\newblock Reinforcement learning with competitive ensembles of
  information-constrained primitives.
\newblock \emph{arXiv preprint arXiv:1906.10667}, 2019.

\bibitem[Holland(1985)]{holland1985properties}
Holland, J.~H.
\newblock Properties of the bucket brigade.
\newblock In \emph{Proceedings of an International Conference on Genetic
  Algorithms and their Applications 1-7}, 1985.

\bibitem[Kingma \& Ba(2014)Kingma and Ba]{kingma2014adam}
Kingma, D.~P. and Ba, J.
\newblock Adam: A method for stochastic optimization.
\newblock \emph{arXiv preprint arXiv:1412.6980}, 2014.

\bibitem[Kwee et~al.(2001)Kwee, Hutter, and Schmidhuber]{kwee2001market}
Kwee, I., Hutter, M., and Schmidhuber, J.
\newblock Market-based reinforcement learning in partially observable worlds.
\newblock In \emph{International Conference on Artificial Neural Networks},
  pp.\  865--873. Springer, 2001.

\bibitem[Mazumdar et~al.(2019)Mazumdar, Ratliff, Jordan, and
  Sastry]{mazumdar2019policy}
Mazumdar, E., Ratliff, L.~J., Jordan, M.~I., and Sastry, S.~S.
\newblock Policy-gradient algorithms have no guarantees of convergence in
  continuous action and state multi-agent settings.
\newblock \emph{arXiv preprint arXiv:1907.03712}, 2019.

\bibitem[Minsky(1988)]{minsky1988society}
Minsky, M.
\newblock \emph{Society of mind}.
\newblock Simon and Schuster, 1988.

\bibitem[Ohsawa et~al.(2018)Ohsawa, Akuzawa, Matsushima, Bezerra, Iwasawa,
  Kajino, Takenaka, and Matsuo]{ohsawa2018neuron}
Ohsawa, S., Akuzawa, K., Matsushima, T., Bezerra, G., Iwasawa, Y., Kajino, H.,
  Takenaka, S., and Matsuo, Y.
\newblock Neuron as an agent, 2018.
\newblock URL \url{https://openreview.net/forum?id=BkfEzz-0-}.

\bibitem[Paszke et~al.(2019)Paszke, Gross, Massa, Lerer, Bradbury, Chanan,
  Killeen, Lin, Gimelshein, Antiga, et~al.]{paszke2019pytorch}
Paszke, A., Gross, S., Massa, F., Lerer, A., Bradbury, J., Chanan, G., Killeen,
  T., Lin, Z., Gimelshein, N., Antiga, L., et~al.
\newblock Pytorch: An imperative style, high-performance deep learning library.
\newblock In \emph{Advances in neural information processing systems}, pp.\
  8026--8037, 2019.

\bibitem[Pathak et~al.(2019)Pathak, Lu, Darrell, Isola, and
  Efros]{pathak2019learning}
Pathak, D., Lu, C., Darrell, T., Isola, P., and Efros, A.~A.
\newblock Learning to control self-assembling morphologies: a study of
  generalization via modularity.
\newblock In \emph{Advances in Neural Information Processing Systems}, pp.\
  2295--2305, 2019.

\bibitem[Peng et~al.(2019)Peng, Chang, Zhang, Abbeel, and Levine]{peng2019mcp}
Peng, X.~B., Chang, M., Zhang, G., Abbeel, P., and Levine, S.
\newblock Mcp: Learning composable hierarchical control with multiplicative
  compositional policies.
\newblock \emph{arXiv preprint arXiv:1905.09808}, 2019.

\bibitem[Plato(380 B.C.)]{plato380republic}
Plato.
\newblock \emph{Republic}.
\newblock 380 B.C.

\bibitem[Rosenbaum et~al.(2017)Rosenbaum, Klinger, and
  Riemer]{rosenbaum2017routing}
Rosenbaum, C., Klinger, T., and Riemer, M.
\newblock Routing networks: Adaptive selection of non-linear functions for
  multi-task learning.
\newblock \emph{arXiv preprint arXiv:1711.01239}, 2017.

\bibitem[Roughgarden(2016)]{roughgarden2016twenty}
Roughgarden, T.
\newblock \emph{Twenty lectures on algorithmic game theory}.
\newblock Cambridge University Press, 2016.

\bibitem[Schmidhuber(1987)]{schmidhuber1987evolutionary}
Schmidhuber, J.
\newblock \emph{Evolutionary principles in self-referential learning, or on
  learning how to learn: the meta-meta-... hook}.
\newblock PhD thesis, Technische Universit{\"a}t M{\"u}nchen, 1987.

\bibitem[Schmidhuber(1989)]{schmidhuber1989local}
Schmidhuber, J.
\newblock A local learning algorithm for dynamic feedforward and recurrent
  networks.
\newblock \emph{Connection Science}, 1\penalty0 (4):\penalty0 403--412, 1989.

\bibitem[Schulman et~al.(2015)Schulman, Moritz, Levine, Jordan, and
  Abbeel]{schulman2015high}
Schulman, J., Moritz, P., Levine, S., Jordan, M., and Abbeel, P.
\newblock High-dimensional continuous control using generalized advantage
  estimation.
\newblock \emph{arXiv preprint arXiv:1506.02438}, 2015.

\bibitem[Schulman et~al.(2017)Schulman, Wolski, Dhariwal, Radford, and
  Klimov]{schulman2017proximal}
Schulman, J., Wolski, F., Dhariwal, P., Radford, A., and Klimov, O.
\newblock Proximal policy optimization algorithms.
\newblock \emph{arXiv preprint arXiv:1707.06347}, 2017.

\bibitem[Selfridge(1988)]{selfridge1988pandemonium}
Selfridge, O.~G.
\newblock \emph{Pandemonium: A Paradigm for Learning}, pp.\  115–122.
\newblock MIT Press, Cambridge, MA, USA, 1988.
\newblock ISBN 0262010976.

\bibitem[Srivastava et~al.(2013)Srivastava, Masci, Kazerounian, Gomez, and
  Schmidhuber]{srivastava2013compete}
Srivastava, R.~K., Masci, J., Kazerounian, S., Gomez, F., and Schmidhuber, J.
\newblock Compete to compute.
\newblock In \emph{Advances in neural information processing systems}, pp.\
  2310--2318, 2013.

\bibitem[Sutton(1988)]{sutton1988learning}
Sutton, R.~S.
\newblock Learning to predict by the methods of temporal differences.
\newblock \emph{Machine learning}, 3\penalty0 (1):\penalty0 9--44, 1988.

\bibitem[Sutton et~al.(1999)Sutton, Precup, and Singh]{sutton1999between}
Sutton, R.~S., Precup, D., and Singh, S.
\newblock Between mdps and semi-mdps: A framework for temporal abstraction in
  reinforcement learning.
\newblock \emph{Artificial intelligence}, 112\penalty0 (1-2):\penalty0
  181--211, 1999.

\bibitem[Vickrey(1961)]{vickrey1961counterspeculation}
Vickrey, W.
\newblock Counterspeculation, auctions, and competitive sealed tenders.
\newblock \emph{The Journal of finance}, 16\penalty0 (1):\penalty0 8--37, 1961.

\end{thebibliography}
\bibliographystyle{icml2020}

\clearpage
\appendix
\onecolumn
\section{Game Theory}
In the context of this paper, a \textbf{strategy} of a primitive $\omega$ is equivalent to its bidding policy $\psi$.
A \textbf{strategy profile} is the set of strategies $\psi^{1:N}$ for all primitives $\omega^{1:N}$.
For emphasis, we equivalently write the utility $\yourutility\left(\psi^{1:N}\right)$ for player $i$ as $\yourutility\left(\psi^i; \psi^{-i}\right)$, where $\psi^i$ is the strategy for player $i$ and $\psi^{-i}$ is the strategy for all other players.

\begin{definition}
\textbf{Best Response}: A strategy $\psi^i$ is the best response for player $i$ if given the strategies of all other players $\psi^{-i}$, $\psi^i$ maximizes player $i$'s utility $\yourutility\left(\psi^i; \psi^{-i}\right)$.
\end{definition}
\begin{definition}
\textbf{Nash Equilibrium}: A strategy profile is in a Nash equilibrium if given the strategies of other players, each player's strategy is a best response.
\end{definition}
\begin{definition}
\textbf{Dominant Strategy Equilibrium}: A strategy $\psi$ is a dominant strategy if it is the best response for a player no matter what strategies the other players play. A dominant strategy equilibrium is the unique Nash equilibrium where every player plays their dominant strategy.
\end{definition}
\begin{definition}
\textbf{Weakly Dominated Strategies}: 
Consider player $\yourindex$ playing strategy $\psi^i$.
Let the strategies for all other players be $\psi^{-\yourindex}$.
A strategy $\psi^i$ \textbf{weakly dominates} $\tilde{\psi}^i$ if for all strategies of other players $\psi^{-\yourindex}$, $\yourutility(\psi; \psi^{-\yourindex}) \geq \yourutility(\tilde{\psi};\psi^{-\yourindex})$, and there exists a $\tilde{\psi}^{-\yourindex}$ such that $\yourutility(\psi^i;\tilde{\psi}^{-\yourindex}) > \yourutility(\tilde{\psi}^i;\tilde{\psi}^{-\yourindex})$. $\psi$ is \textbf{dominated} if there exists a strategy which dominates it.
\end{definition}
\begin{definition}
\textbf{Iterated Deletion of Dominated Strategies}: 
Iterated deletion of dominated strategies repeatedly (a) removes all dominated strategies for each player, then (b) updates the dominance relation (as now there are fewer strategies of other players to consider). It terminates when all remaining strategies are undominated.
\end{definition}

\paragraph{Vickrey Auction}
The Vickrey Auction is a type of single-item sealed bid auction.
It is single-item, which means there is a single auction item up for sale.
It is sealed-bid, which means that the players have no knowledge of each others' bids.

\section{Societal Decision-Making Framework}
{\renewcommand{\arraystretch}{2.0}
\begin{table}[h]
    \centering
    \begin{tabular}{ c|c|c } 
        \hline
        Abstraction Level & Global & Local \\ 
        \hline
        \hline
        Agent & Society $\Omega = \{\omega^i\}_{i=1}^N$ & Primitive $\omega^i = \left(\psi^i, \phi^i\right)$ \\ 
        \arrayrulecolor{gray}\hline
        Environment & global MDP & local auction at state $s$ \\ 
        State Space & $\mathcal{S}$ & the single state $\{s\}$ \\ 
        Action Space & the indices of the primitives $\mathcal{A} = \{1, 2, ..., N\}$ & the space of non-negative bids $\mathcal{B} = \mathbb{R}_{\geq 0}$ \\
        \arrayrulecolor{gray}\hline
        Objective & $J(\pi_\Omega) = \mathbb{E}_{\tau \sim p^\pi(\tau)}\left[\sum_{t=0}^T \gamma^t r\left(s_t, \omega_t\right)\right]$ & $\yourutility_s(\allbidders) = \yourvalue_s \cdot \yourallocation(\mathbf{b}) - \yourpayment(\mathbf{b})$ \\ 
        \arrayrulecolor{gray}\hline
        Problem & $\max_{\pi_\Omega} J(\pi_\Omega)$ & $\max_{\psi^i} \yourutility_s(\allbidders)$ \\
        \hline
    \end{tabular}
    \caption{\textbf{Societal Decision-Making.} This table specifies the \textbf{agent}, \textbf{environment}, \textbf{objective}, and \textbf{problem} at both the \textbf{global} and \textbf{local} levels of abstraction in the societal-decision-making framework.}
    \label{tab:societal_decision_making}
\end{table}}
Though both the monolithic decision-making framework as well as the societal decision-making (in which the transformations $\phi$ correspond to literal actions) can both be used for reinforcement learning, the key difference between the two is that the learnable parameters are trained to optimize the objective of the MDP in the monolithic framework, whereas the learnable parameters are trained to optimize the objective of the auction at each state of the MDP in the societal framework.

\section{The Cloned Vickrey Society as a Solution to Societal Decision-Making}

The MDP specifies the environment.
The society specifies the abstract agent that interacts in the environment.
The \textbf{Market MDP} is a global MDP governed by a specific type of auction mechanism (the Vickrey mechanism) that is agnostic to the architecture of the society that interacts with it.
A \textbf{cloned society} is a specific architecture of society (one with redundant primitives) that is agnostic to the auction mechanism governing the global MDP. 
The \textbf{cloned Vickrey society} specifies a specific architecture of a society (a cloned society) that interacts with a specific type of global MDP (a Market MDP).

\subsection{Proofs}

\textbf{Proposition~\ref{proposition:optimal_policy}.} 
\textit{Assume at each state $s$ the local auction allocates $\yourallocation(\mathbf{b})=1$ if $i$ wins and $\yourallocation(\mathbf{b})=0$ if $i$ loses.
Then all primitives $\yourprimitive$ bidding their optimal societal Q-values $\societalQ^*(s, \yourprimitive)$ collectively induce an optimal global policy.}
\begin{proof}
For state $s$, the index of the primitive with the highest bid is equal to the primitive with the highest optimal societal Q-value for that state.
That is, $\argmax_\yourindex \mathbf{b}^\yourindex_t = \argmax_\yourindex \societalQ^*(s_t, \primitive_t^\yourindex)$.
Thus, selecting the highest-bidding primitive to win by definition follows the optimal policy for the Market MDP.
\end{proof}

\textbf{Theorem~\ref{thm:dominant_strategy_equilibrium_Q}}
\textit{
If the valuations $\yourvalue_s$ for each state $s$ are the optimal societal Q-values $\societalQ^*(s, \yourprimitive)$, then the society's optimal global policy coincides with the primitives' unique dominant strategy equilibrium under the Vickrey mechanism.
}
\begin{proof}
By defining the valuation of each primitive $\yourprimitive$ to be its optimal societal $Q$-value $\societalQ^*(s_t, \primitive_t^\yourindex)$, the DSIC property of the Vickrey auction guarantees the dominant strategy equilibrium is to bid exactly $\societalQ^*(s_t, \primitive_t^\yourindex)$.
The welfare-maximization property of the Vickrey auction guarantees if all primitives played their dominant strategies, then the primitive with the highest valuation wins, so specifying the valuations such that activating the primitive with the highest valuation at time $t$ follows the optimal global policy at that timestep.
\end{proof}

\textbf{Proposition~\ref{proposition:nash_equilibrium}.}
\textit{
In a Market MDP, it is a Nash equilibrium for every primitive to bid $\societalQ^*(s, \yourprimitive)$.
Moreover, if the Market MDP is finite horizon, then bidding $\societalQ^*(s, \yourprimitive)$ is the unique Nash equilibrium that survives iterated deletion of weakly dominated strategies.
}
\begin{proof}
Consider time-step $t$.
Assuming that every other primitive at every other time-step bids $\societalQ^*$, the best response under the Vickrey mechanism for primitive $\yourprimitive$ at timestep $t$ would be to truthfully bid $r(s_t, \yourprimitive_t) + \gamma \cdot \max_\nextindex \, \societalQ^*(s_{t+1}, \primitive^k)$,
which by definition of the Bellman optimality equations~\citep{bellmann1957dynamic} is $\societalQ^*(s_t, \yourprimitive_t)$.

For the finite horizon case, we proceed by backwards induction.

\textbf{Base Case}: The last time-step $T$ is a bandit problem where $\societalQ^*(s_T, \yourprimitive_T) = r(s_T, \yourprimitive_T)$, so by Theorem~\ref{thm:dominant_strategy_equilibrium_Q} bidding $\societalQ^*(s_t, \yourprimitive_t)$ is the unique dominant strategy.

\textbf{Inductive Hypothesis}: In time-step $t+1$, the strategy which survives iterated deletion of dominated strategies is to bid the optimal societal Q-value.

\textbf{Inductive Step}: By the Inductive Hypothesis, in time-step $t+1$, all primitives bid their optimal societal Q-values if they use any strategy which survives iterated deletion of dominated strategies. 
This means that in time-step $t$, each primitive's valuation for winning is given exactly by their optimal societal Q-value: $r(s_t, \yourprimitive_t) + \gamma \cdot \max_\nextindex \, \societalQ^*(s_{t+1}, \primitive^k)$. 
Therefore, it is a dominant strategy to bid $\societalQ^*$, and all other bids are dominated (and therefore removed).
\end{proof}

\textbf{Lemma~\ref{lemma:credit_conservation}}
\textit{
For a cloned society, at the Nash equilibrium specified in Proposition~\ref{proposition:nash_equilibrium}, what the winning primitive $\winnerprimitive^i$ at time $t$ receives from the winning primitive $\winnerprimitive^k$ at $t+1$ is exactly what $\winnerprimitive^k$ pays: $\mathbf{b}^k_{s_{t+1}}$.
}
\begin{proof}
If two primitives have the same $\phi$, then their societal $Q$-values are identical, so their optimal strategies $\psi$ are identical.
Then at the Nash equilibrium specified in Proposition~\ref{proposition:nash_equilibrium}, their bids are also identical.
\end{proof}

\textbf{Theorem~\ref{theorem:credit_optimal_equilibrium}.}
\textit{
Define a \textbf{cloned Vickrey society} as a cloned society that solves a Market MDP.
Then it is a Nash equilibrium for every primitive in the cloned Vickrey society to bid $\societalQ^*(s, \yourprimitive)$.
In addition, the price that the winning primitive pays for winning is equivalent to what it bid.
}
\begin{proof}
(1) is by Proposition~\ref{proposition:nash_equilibrium} and (2) is by
Lemma~\ref{lemma:credit_conservation}.
\end{proof}

\section{Decentralized Reinforcement Learning Algorithms for the Cloned Vickrey Society} \label{appdx:decentralized_rl}
Section~\ref{sec:mechanism} presented the cloned Vickrey society as a concrete instantiation of the societal decision-making framework whose optimal global policy emerges as a Nash equilibrium of self-interested primitives engaging in local economic transactions.
Section~\ref{sec:learning} removed the assumption that primitives know their valuations and presented decentralized RL algorithms for learning these valuations through interaction.

In this paper, we specified the class of decentralized RL algorithms by the learning objective -- the set of local auction utilities at every state.
We present in Algorithm~\ref{alg:multi_agent_societal_rl} the algorithm pseudocode for an on-policy variant of such a decentralized RL algorithm, but the learning objective is in principle agnostic to the RL algorithm use to train each primitive.
However, simply specifying only one variant within this class of algorithm in this paper leaves much to still be explored.
Adapting methods developed in the monolithic framework, including bandit algorithms, off-policy algorithms, and on-policy algorithms, for optimizing the local auction utilities the would be an interesting direction for future work.

\begin{algorithm}[H]
    \caption{On-Policy Decentralized RL}
    \label{alg:multi_agent_societal_rl}
    \small
    \begin{algorithmic}[1]
        \State \textbf{Initialize:} Primitives $\primitive^{1:\lastagentindex}$, Memory $m^{1:\lastagentindex}$, RL update rule $f$
        \While{True}
        \LineComment{\textit{Sample Episode}}
        \While{episode has not terminated}
            \State $\primitive^{1:\lastagentindex}$ observe state $s$
            \State $\primitive^{1:\lastagentindex}$ produce bids $\mathbf{b}^1_{s}, ..., \mathbf{b}^\lastagentindex_{s}$
            \State Auction selects winner $\winnerprimitive$ with transformation $\hat{\phi}_{\mathcal{T}}$
            \State $\winnerprimitive$ produces $s' = \hat{\phi}_{\mathcal{T}}(s)$
            \State Record environment reward $r(s, \winnerprimitive)$
            \State $s \gets s'$
        \EndWhile
        \LineComment{\textit{Compute Utilities}}
        \For{each time-step $t$ until the end of sampled episode}
            \State $\winnerprimitive_t$ gets $\winnerutility_{s_t}(\allprimitives) = r(s_t, \winnerprimitive_t) + \gamma \cdot \underset{k}{\max} \mathbf{b}^k_{s_{t+1}} - \underset{j \neq i}{\max} \mathbf{b}^j_{s_t}$
            \State Losers $\omega^j_t$ get $U_{s_t}^j(\allprimitives) = 0$
            \For{all primitives $\yourprimitive$}
                \State Primitive $\yourprimitive_t$ stores $(s_t, \mathbf{b}^i_{s_t}, s_{t+1}, \yourutility_{s_t}(\yourprimitive_t))$ into $m^i$
            \EndFor
        \EndFor
        \LineComment{\textit{Update}}
        \If{time to update}
             \For{all primitives $\yourprimitive$}
                \State Update primitive $\yourprimitive$ with update rule $f$ with memory $m^i$
             \EndFor
        \EndIf
        \EndWhile
    \end{algorithmic}
\end{algorithm}

\section{Implementations of an On-Policy Decentralized Reinforcement Learning Algorithm}
\paragraph{Stochastic policy gradient}
In this paper, we considered optimizing the bidding policies with a stochastic policy gradient algorithm, which means that the bidding policies parameterize a distribution of bids.
The stochasticity of the bidding policy means that there need not be an explicit exploration strategy such as $\epsilon$-greedy, which simplifies the analysis in this paper.
In future work it would be interesting to explore deterministic bidding policies as well.

\paragraph{Bidding Policies}
We implement all bidding policies as neural networks that output the parameters of a Beta-distribution.
Because the Beta-distribution has support between 0 and 1, we normalized environment rewards so returns fit in this range.
In theory, the range of possible bids could be $[0, \infty)$ and need not be restricted to $[0, 1]$.
The society decision-making framework prescribes a different unique local auction, with different primitives, at each state $s$.
However, because the state space could be extremely large or even continuous, in practice we share the same set of primitives $\omega^{1:N}$ across all states, express their bidding policies as functions of the state, and rely on the function approximation capabilities of neural networks to learn different state-conditioned bidding strategies.

\paragraph{Redundancy}
Implementing two clones of each primitive means that each clone should share the same transformation $\phi$: if $\phi$ were the literal action of \texttt{go-left}, then there would be two primitives that bid in the local auction to execute the \texttt{go-left} action.
We implemented redundant clones by sharing the weights of their bidding policies $\psi$.
Therefore, cloned primitives have identical bidding distributions, from which different bids are sampled.
Alternatively, we also explored giving clone primitives the same transformation $\phi$ but independently parameterized bidding policies $\psi$.
While we did not find much difference in global performance with independently parameterized bidding policies, such a scheme could be useful for multi-task learning where the same transformation could have different optimal societal Q-values depending on the task.

\paragraph{Learning Objectives}
The utility of the cloned Vickrey society is the learning objective.
We compared three possible implementations of this learning objective: bucket-brigade (\textit{BB}), Vickrey (\textit{V}), and credit-conserving Vickrey (\textit{CCV}) as described in Section~\ref{sec:empirical_mdps} and Figure~\ref{fig:possible_implementations}.
We also compared with a baseline that sets the learning objective to be the environment reward.
For all implementations and the baseline, the learning objective that a loser $\primitive$ that the auction receives is $0$ because its allocation $\yourallocation(\mathbf{b})$ and payment $\yourpayment(\mathbf{b})$ are both $0$.
Letting $\hat{\mathbf{b}}_{s_t}$ and $\mathbf{b}'_{s_t}$ denote the highest and second highest bid at time $t$ respectively, the learning objective $\hat{U}^i_{s_t}(\allprimitives)$ of the winner $\winnerprimitive$ for state $s_t$ is given below.

{\renewcommand{\arraystretch}{2.0}
\begin{table}[h]
    \centering
    \begin{tabular}{ c|c } 
        \hline
         & Learning Objective \\ 
        \hline
        \hline
        Bucket-Brigade & $\winnerutility_{s_t}(\allprimitives) = \left[r(s_t, \winnerprimitive_t) + \gamma \cdot \hat{\mathbf{b}}_{s_{t+1}}\right] - \hat{\mathbf{b}}_{s_t}$ \\ 
        Vickrey & $\winnerutility_{s_t}(\allprimitives) = \left[r(s_t, \winnerprimitive_t) + \gamma \cdot \hat{\mathbf{b}}_{s_{t+1}}\right] - \mathbf{b}'_{s_t}$ \\ 
        Credit-Conserving Vickrey & $\winnerutility_{s_t}(\allprimitives) = \left[r(s_t, \winnerprimitive_t) + \gamma \cdot \mathbf{b}'_{s_{t+1}}\right] - \mathbf{b}'_{s_t}$ \\ 
        \arrayrulecolor{gray}\hline
        Environment Reward & $\winnerutility_{s_t}(\allprimitives) = \left[r(s_t, \winnerprimitive_t)\right]$ \\ 
        \hline
    \end{tabular}
\end{table}}

\section{Experimental Details}
We implemented our experiments using the PyTorch library~\citep{paszke2019pytorch}.
For all experiments, for proximal policy optimization we used a policy learning rate of $4\cdot 10^{-5}$, a value function learning rate of $5 \cdot 10^{-3}$, a clipping ratio of 0.2, a GAE~\citep{schulman2015high} smoothing parameter of 0.95, a discount factor of 0.99, and the Adam~\citep{kingma2014adam} optimizer.
Each bidding policy has its own replay buffer.
Each bidding policy is updated every 4096 transitions with a minibatch size of 256.
For \textit{Two Rooms} and \textit{Mental Rotation} we added an entropy bonus, with a weighting coefficient of $0.1$, to the PPO loss.

The plots for \textit{Market Bandit}, \textit{Two Rooms}, and \textit{Mental Rotation} were averaged over 3 seeds and the other plots were averaged over 5 seeds.
The metric for the learning curves is the mean return over 4096 environment steps.
The error bars represent the 10th, 50th, 90th quantile.
All experiments were run on CPU, except for the \textit{TwoRooms} environment, which was run on GPU.

\subsection{Numerical Simulations}
All transformations $\phi$ correspond to literal actions.
The bidding policies $\psi$ are implemented as linear neural networks with a single hidden layer of 16 units that output the $\alpha$ and $\beta$ parameters of the Beta distribution.
There is no activation function for the hidden layer.
We used the softplus activation function to output $\alpha$ and $\beta$.
The valuation functions for the bidding policies are also implemented as linear neural networks with a single hidden layer of 16 units.

\subsubsection{Market Bandit}
At every round, we stochastically drop out a subset of primitives.
For cloned societies, one clone could be dropped out while the other clone stays in the auction.
The drop-out sampling procedure is as follows.
Letting $N$ be the number of total primitives (which means there are $N/2$ unique primitives in cloned societies), we first sample a random integer $m$ in $\{2, 3, ..., N\}$.
Then we sample a subset of $m$ primitives from the $N$ total primitives without replacement.
For a given round, only the primitives that participated in that round are updated.

\subsubsection{Duality} \label{appdx:duality_details}
\begin{figure}[h]
    \centering
    \includegraphics[width=0.49\textwidth]{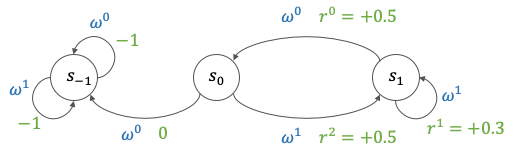}
    \caption{\small{\textbf{The \textit{Duality} environment.}
    }}
    \label{fig:duality}
\end{figure}
As stated in Figure~\ref{fig:possible_implementations}, without redundant primitives the solitary \textit{CCV} implementation sacrifices Bellman optimality in general.
We now use the \textit{Duality} environment as an example to illustrate such an instance where the dominant strategy of the primitives does not coincide with the global optimal policy of the society, even if the primitives have full knowledge of their own valuations.
The DSIC property of the Vickrey auction makes it straightforward to analyze the dominant strategies of the primitives, which the following paragraphs illustrate.
Recall that based on the \textit{CCV} learning objective, the valuation of the winner $\winnerprimitive$ at time $t$ is the immediate reward plus the discounted \textit{second}-highest bid (not the highest bid) at the next time-step: $r(s_t, \winnerprimitive_t) + \gamma \cdot \mathbf{b}'_{s_{t+1}}$.
Let $r^0$ be equal to the environment reward $r(s_1, \primitive^0)$, $r^1$ be equal to the environment reward $r(s_1, \primitive^1)$, and $r^2$ be equal to the environment reward $r(s_0, \primitive^1)$, where in Figure~\ref{fig:duality} we see that $r^0 = 0.5$, $r^1 = 0.3$, and $r^2 = 0.5$

At state $s_0$, primitive $\omega^0$ will bid $0$, the lowest possible bid, because the local auction at state $s_0$ would lead to unbounded negative reward at $s_{-1}$.
This means that the valuation that $\omega^0$ has for winning at state $s_1$ is $r^0+ \gamma \cdot 0 = r(s_1, \primitive^0)$, since $0$ must be the second highest bid at state $s_0$.
Thus the dominant strategy for $\omega^0$ is to truthfully bid $r^0$.
At $s_1$, primitive $\omega^1$ will bid some number $c$.
Since activating $\omega^1$ is a self-loop, $\omega^1$ can sell $s_1$ back to itself in the next time-step.
Thus the valuation that $\omega^1$ has for winning at state $s_1$ is $r^1 + \gamma \cdot \min(r^0, c)$.
Thus the dominant strategy for $\omega^0$ is to truthfully bid $c = r^1 + \gamma \cdot \min(r^0, c)$.

In the undiscounted case, where $\gamma = 1$, if we solve for $c$, we have that if $r^1 > 0$, then $c = r^1 + r^0$, which is greater than $r^0$, which is what $\omega^0$ would bid as its dominant strategy.
Therefore, in the case that $r^1 > 0$, $\omega^1$ will continue to sell $s_1$ to itself.
As long as $r^1 < r^0 + r^2$, the self-loop at $s_1$ will be less optimal than cycling back and forth between $s_0$ and $s_1$.

In the discounted case, where $0 < \gamma < 1$ and we again assume that $r^1 > 0$.
Solving $c$ again, we see that if in the case that $r^0 < \frac{r^1}{1-\gamma}$, then $c = r^1 + \gamma r^0$ and $c > r^0$.
In this case $\omega^1$ will bid higher than $\omega^0$ at state $s_1$, creating a perpetual self-loop.
If instead $r^0 > \frac{r^1}{1-\gamma}$, then $c = \frac{r^1}{1-\gamma}$ and $c < r^0$.
In this case $\omega^0$ will bid higher than $\omega^1$ at state $s_1$, which is the optimal global policy.

Therefore the \textit{Duality} environment illustrates that without redundancy, for fortuitous settings of $r^0$ and $r^1$, the dominant strategy equilibrium of the society may coincide with the optimal global policy, but if the for other choices of $r^0$ and $r^1$, the dominant strategy equilibrium of the society may be globally suboptimal.
Adding redundant primitives makes the auction utilities consistent with the Bellman optimality equations and therefore does not suffer from such suboptimal equilibria.

\subsection{Two Rooms}
\begin{figure}[h]
    \centering
    \includegraphics[width=0.49\textwidth]{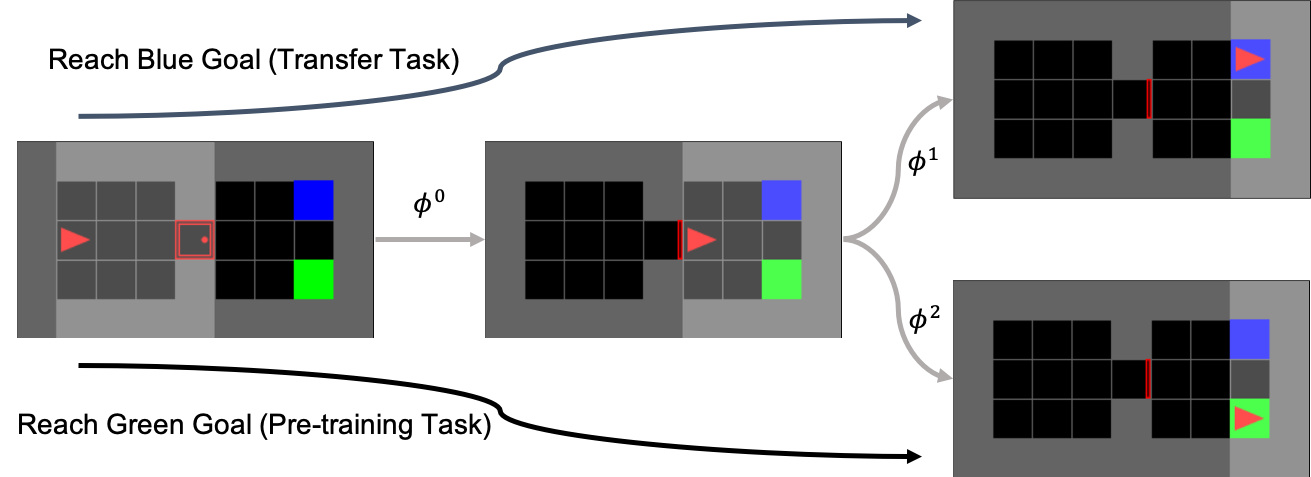}
    \caption{\small{\textbf{The \textit{Two Rooms} environment.}
    }}
    \label{fig:hrl_appendix}
\end{figure}
The transformations $\phi^{0:2}$ are subpolicies pre-trained with PPO and have an action space equivalent to the action space of the Minigrid environment.
For these transformation subpolicies, the bidding policies, the value functions for the bidding policies, the non-hierarchical monolithic baseline, and the hierarchical monolithic baseline, we adapted the convolutional neural network architecture from~\url{https://github.com/lcswillems/rl-starter-files}.

$\phi^0$ is initialized randomly in the room on the left and is pre-trained to take the \text{done} action once it opens the red door and enters the room on the right, upon which it receives a terminal reward.
$\phi^1$ is initialized randomly in the room on the right and is pre-trained to take the \text{done} action upon reaching the green square, upon which it receives a terminal reward.
$\phi^2$ is initialized randomly in the room on the right and is pre-trained to take the \text{done} action upon reaching the blue square, upon which it receives a terminal reward.

In the pre-training task, the society receives a terminal reward upon reaching the green square and no intermediate rewards.
In the transfer task, the society receives a terminal reward upon reaching the blue square and no intermediate rewards.
These subpolicies $\phi^{0:2}$ are frozen for these tasks.
For all learners, we trained to convergence on the pre-training task, then we initialized training for the transfer task from the best saved checkpoint on the pre-training task.
The non-hierarchical monolithic baseline could not solve the pre-training task so we did not consider it for the transfer task.

\subsection{Mental Rotation}
\begin{figure}[h]
    \centering
    \includegraphics[width=0.7\textwidth]{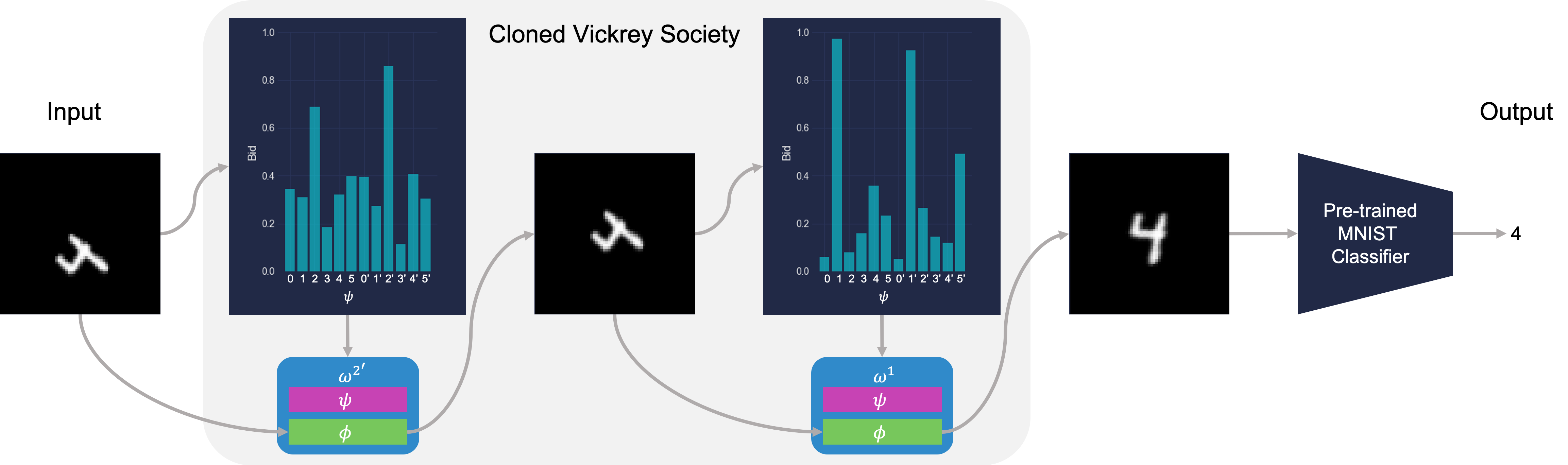}
    \caption{\small{\textbf{The \textit{Mental Rotation} environment.}
    }}
    \label{fig:mnist_appendix}
\end{figure}
\paragraph{Environment}
Each $28 \times 28$ image in the MNIST training set was first inset into a black background of $64 \times 64$.
Then the image first rotated either clockwise or counterclockwise by 60 degrees, then translated left, up, down, right by 29\% of the image width, for a total of eight possible transformation combinations given these six affine transformations.
These transformation combinations were taken from~\citet{chang2018automatically}.

\paragraph{Primitives} 
The transformations $\phi$ correspond to the same six affine transformations, summarized below:
{\renewcommand{\arraystretch}{2.0}
\begin{table}[h]
    \centering
    \begin{tabular}{ c|c } 
        \hline
        $\phi^0$, ${\phi^0}'$ & \texttt{rotate-counterclockwise} \\ 
        $\phi^1$, ${\phi^1}'$ & \texttt{rotate-clockwise} \\ 
        $\phi^2$, ${\phi^2}'$ & \texttt{translate-up} \\ 
        $\phi^3$, ${\phi^3}'$ & \texttt{translate-down} \\ 
        $\phi^4$, ${\phi^4}'$ & \texttt{translate-left} \\ 
        $\phi^5$, ${\phi^5}'$ & \texttt{translate-right} \\ 
        \hline
    \end{tabular}
\end{table}}

The primitive $\omega^i$, and its bidding policy $\psi^i$, correspond to the transformation $\phi^i$.
The clones are indicate by $i$ and $i'$.

\paragraph{Neural network architecture}
The pre-trained MNIST classifier, the bidding policies, and the value functions for the bidding policies follow the same architecture as the convolutional neural network used in~\citet{chang2018automatically}, with different output dimensions ($10$ as the output dimension of the MNIST classifier, $1$ for the other networks).
The PyTorch architecture is given below:
\begin{verbatim}
network = nn.Sequential(
            nn.Conv2d(1, 8, 4, 2, 1, bias=False),
            nn.LeakyReLU(0.2, inplace=True),
            nn.Conv2d(8, 8 * 2, 4, 2, 1, bias=False),
            nn.BatchNorm2d(8 * 2),
            nn.LeakyReLU(0.2, inplace=True),
            nn.Conv2d(8 * 2, 8 * 4, 4, 2, 1, bias=False),
            nn.BatchNorm2d(8 * 4),
            nn.LeakyReLU(0.2, inplace=True),
            nn.Conv2d(8 * 4, 8 * 8, 4, 2, 1, bias=False),
            nn.BatchNorm2d(8 * 8),
            nn.LeakyReLU(0.2, inplace=True),
            nn.Conv2d(8 * 8, outdim, 4, 1, 0, bias=False))
\end{verbatim}

\end{document}